\documentclass[10pt,twocolumn,letterpaper]{article}

\usepackage[pagenumbers]{cvpr} 

\usepackage{graphicx}
\usepackage{amsmath}
\usepackage{amssymb}
\usepackage{booktabs}
\usepackage{times}
\usepackage{epsfig}
\usepackage{amsthm}
\usepackage{bm}
\usepackage{xcolor}
\usepackage{amsfonts}
\usepackage{multirow}
\usepackage{threeparttable}
\usepackage{float}
\usepackage{adjustbox}
\usepackage[ruled,linesnumbered]{algorithm2e}
\usepackage{colortbl}

\usepackage[pagebackref,breaklinks,colorlinks]{hyperref}

\usepackage[capitalize]{cleveref}
\crefname{section}{Sec.}{Secs.}
\Crefname{section}{Section}{Sections}
\Crefname{table}{Table}{Tables}
\crefname{table}{Tab.}{Tabs.}

\definecolor{commentcolor}{RGB}{110,154,155}  
\newcommand{\PyComment}[1]{\ttfamily\textcolor{commentcolor}{\# #1}}  
\newcommand{\PyCode}[1]{\ttfamily\textcolor{black}{#1}}

\def\cvprPaperID{*****} 
\def\confName{CVPR}
\def\confYear{2024}

\begin{document}

\title{Language-Guided Diffusion Model for Visual Grounding}

\author{Sijia Chen, Baochun Li\\
\textit{Department of Electrical and Computer Engineering}\\
\textit{University of Toronto}\\
{\tt\small sjia.chen@mail.utoronto.ca, \tt\small bli@ece.toronto.edu}
}

\maketitle

\begin{abstract}
    Visual grounding (VG) tasks involve explicit cross-modal alignment, as semantically corresponding image regions are to be located for the language phrases provided. Existing approaches complete such visual-text reasoning in a single-step manner. Their performance causes high demands on large-scale anchors and over-designed multi-modal fusion modules based on human priors, leading to complicated frameworks that may be difficult to train and overfit to specific scenarios. Even worse, such once-for-all reasoning mechanisms are incapable of refining boxes continuously to enhance query-region matching. In contrast, in this paper, we formulate an iterative reasoning process by denoising diffusion modeling. Specifically, we propose a language-guided diffusion framework for visual grounding, LG-DVG, which trains the model to progressively reason queried object boxes by denoising a set of noisy boxes with the language guide. To achieve this, LG-DVG gradually perturbs query-aligned ground truth boxes to noisy ones and reverses this process step by step, conditional on query semantics. Extensive experiments for our proposed framework on five widely used datasets validate the superior performance of solving visual grounding, a cross-modal alignment task, in a generative way. The source codes are available at \url{https://github.com/iQua/vgbase/tree/main/examples/DiffusionVG}. 
\end{abstract}

\section{Introduction}

Given the text query and image, visual grounding aims to reason image regions that are semantically aligned with language phrases. To this end, two major categories of visual grounding can be identified: phrase localization \cite{Flickr30kE, TwoBranch} and referring expression comprehension \cite{RIG, MContextRE}, which are essential techniques for bridging the gap between linguistic expressions and visual perception.  

Although extensive frameworks have been proposed to accomplish the two tasks, their fundamental learning principle falls under the once-for-all reasoning mechanism, wherein visual-linguistic alignment is conducted only once at the end of the learning pipeline. This generally results in excessive dependence on the performance of the preceding modules to guarantee effective final reasoning. Consequently, such an over-reliance causes these intermediate modules to become overly complicated, which exacerbates framework issues related to hard-to-train and overfitting to specific scenarios. Specifically, the two-stage framework \cite{CMContextGraph, VSGraph, SeqGROUND, LanguageCues} is limited by heavily relying on region proposals from off-the-shelf detectors. Meanwhile, the one-stage framework \cite{FAOA, RealTimeCMC, TransVG} typically includes sophisticated multi-modal fusion and relation learning modules to merge text and pre-defined dense visual anchors for the final box regression.

\begin{figure}[t]
    \centering
    \includegraphics[width=\columnwidth]{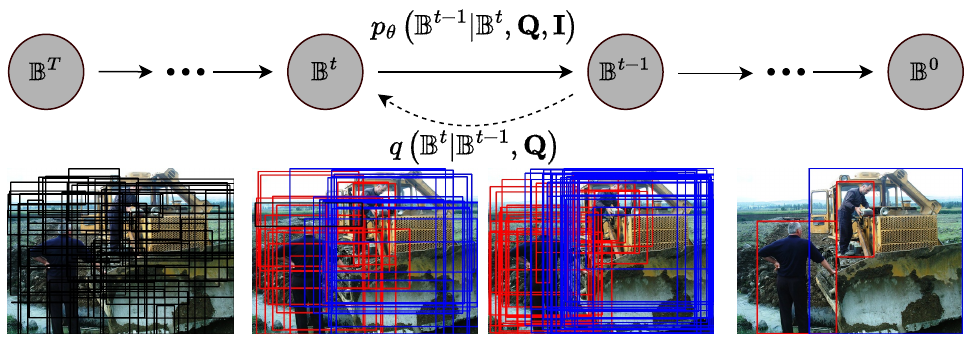}
    \caption{Our proposed visual grounding framework with an iterative learning pipeline built upon the Markov Chain of diffusion models. The forward diffusion process $q$ diffuses query-aligned boxes from ground truth boxes $\mathbb{B}^0$ to ones $\mathbb{B}^T$ with random distribution, while $p$ is the reverse denoising process. Each reverse step relies on a trainable model $\theta$ to make predictions conditional on the query sentence $\bm{Q}$ - "\textcolor{red}{Two men} looking at \textcolor{blue}{a bulldozer}." \textcolor{red}{Two men}" corresponds to two ground truth boxes.}
    \label{fig:LGDVGIntro}
    \vspace{-0.3cm}
\end{figure}

Furthermore, because this sequence learning with once-for-all reasoning lacks the sustainable refinement property, errors induced by any intermediate modules can be subsequently accumulated and amplified to lead to eventual failure. The initial objective of this paper is to address these issues by reconsidering visual grounding via iterative learning \cite{IterativeVisualReasoning, IterativeMatching}, in which visual-text correspondences are not obtained in a one-step manner but through continuous learning and reasoning. 

While the latest works \cite{RecursiveSubQuery, IterativeSRL, VLTVG, MDGT}  that explore such an intuitive notion have demonstrated the notable benefits of iterative reasoning in visual grounding, their framework, especially M-DGT \cite{MDGT} and VLTVG \cite{VLTVG}, still entails intricate modules and a multi-stage iterative process manually designed based on prior human knowledge. Consequently, despite various efforts in this direction, these approaches remain mired in the problems of hard-to-train and overfitting. Moreover, most of them \cite{RecursiveSubQuery, IterativeSRL, VLTVG}, which produce a sole box regression, are inadequate for addressing the one-to-many challenge brought by \texttt{Flickr30K Entities} dataset \cite{Flickr30kE}, wherein one noun-phrase query corresponds to multiple image regions, as shown n Figure~\ref{fig:LGDVGIntro}.

In this work, we propose a \underline{\textbf{l}}anguage-\underline{\textbf{g}}uided \underline{\textbf{d}}iffusion model for \underline{\textbf{v}}isual \underline{\textbf{g}}rounding (\textit{LG-DVG}), thereby mathematically formulating iterative reasoning as a Morkov Chain \cite{BasicDiffusion}. Our language-guided noise-to-box pipeline, shown in Figure \ref{fig:LGDVGIntro} is capable of progressively refining noisy boxes conditional on both boxes attained in the previous step and the text query. Without introducing parameters, the forward diffusion process gradually adds Gaussian noise to query-aligned ground truth boxes step by step till generating noisy boxes with random distribution. Thus, target boxes corresponding to input queries can be recovered by reversing this learned diffusion process with guidance from language semantics. Endowed with a theoretical guarantee from diffusion models \cite{BasicDiffusion, DiffusionBeatGANs}, \textit{LG-DVG} achieves this learning mechanism by training a grounding decoder, comprising a novel cross-modal transformer and a query-conditioned predictor, to reconstruct ground truth boxes from noisy boxes at different time steps.

Built upon diffusion-based iterative learning, \textit{LG-DVG} gains four appealing advantages simultaneously. \emph{First}, naturally performing iterative reasoning in a Markov chain releases the learning framework from handcrafted multi-stage structures. \emph{Second}, the inference stage adjusts a noisy distribution over center coordinates, widths, and heights to query-dependent bounding boxes, thus eliminating any dependence on prepared region candidates or anchors. \emph{Third}, once more inference steps are performed, the progressive refinement property contributes to tighter grounding boxes. Last but not least, \textit{LG-DVG} can, in parallel, adjust several initial bounding boxes with noisy distributions to targets for a query, thereby addressing the one-to-many challenge as described in Figure~\ref{fig:LGDVGIntro}.

To summarize, our contributions are three-fold: 1). By building iterative learning in the context of diffusion models, \textit{LG-DVG} is the first work that models visual-linguistic alignment task, in a generative way for continuous box refinement. 2). Thanks to the proposed language-guided grounding decoder, noisy boxes are able to be adjusted conditional on language semantics to approach ground truth boxes progressively. 3). Experiments on five benchmark visual grounding datasets, including \texttt{Flickr30K Entities} \cite{Flickr30kE}, \texttt{ReferItGame} \cite{Referitgame}, and \texttt{RefCOCO-related} \cite{RefCOCO, RefCOCOg} datasets, comprehensively support our argument on the above benefits. Remarkably, \textit{LG-DVG} achieves competitive accuracy, such as $79.92$ obtained on the test set of \texttt{Flickr30K Entities}, yet with less inference time cost as compared with state-of-the-art competitors.

\section{Related Work}
\subsection{Visual Grounding}
There are mainly two threads of learning frameworks in visual grounding. Those once-for-all reasoning approaches generally embrace two-stage \cite{TwoBranch, MContextRE, Flickr30kE, LanguageCues, SeqGROUND, VSGraph, CMContextGraph} or one-stage \cite{FAOA, RealTimeCMC, TransVG} frameworks. Two-stage ones exploit off-the-shelf detectors \cite{FasterRCNN, DETR} to prepare region candidates from the image, which are subsequently matched with text queries in a ranking structure. Recent progress made by one-stage frameworks avoids the reliance on pre-trained detectors, thus mitigating the efficiency degradation and the adverse effects of undetected objects on subsequent learning. The core idea is to fuse the visual-text feature densely at all spatial locations and directly predict bounding boxes to ground the target.

To break performance limitations often attributed to only reasoning once, the efforts \cite{RecursiveSubQuery, IterativeSRL, VLTVG, MDGT} most related to our idea propose effective frameworks containing a multi-stage reasoning architecture. Specifically, a recursive sub-query construction framework \cite{RecursiveSubQuery} is proposed to reason between image and query for multiple rounds. Emphasizing the importance of reasoning bounding boxes for queries progressively, the work \cite{IterativeSRL} implements an iterative shrinking mechanism to localize the target with reinforcement learning. Recently, VLTVG \cite{VLTVG} develops a multi-stage cross-modal decoder to achieve iterative reasoning. However, the efficacy of these methodologies depends on human priors to build a complex learning process containing multiple manually designed large-scale modules. For instance, although M-DGT \cite{MDGT} achieves box regression recursively, its authors conclude that the designed model is difficult to train. We argue that this may be attributed to the heavy reliance on a multi-stage and tricky learning pipeline, not to mention the intricate node and graph transformers. Our paper proposes to address these issues by extending the Markov Chain to model iterative reasoning with diffusion models in a generative way.

\subsection{Diffusion Model}

Diffusion models \cite{BasicDiffusion, ScoreDiffusion}, a class of probabilistic generative models inspired by nonequilibrium thermodynamics, contain a gradual denoising process, enabling it to convert samples from a random noise distribution to representative data samples. Its remarkable success in natural language processing \cite{StructuredDenoising, DiffusionLM} and image generation \cite{BasicDiffusion, DiffusionBeatGANs} encourages many recent works proposed to extend diffusion models to cross-modal generation tasks such as text-to-image generation \cite{Imagen, GLIDE, DALLE2}, in which text guides how an image is to be generated. This motivates us to investigate the possible benefits of using diffusion models in visual grounding, a learning task involving text and image. More pioneer works are proposed to expose the potential of diffusion models in perception tasks by adopting the diffusion learning paradigm for image segmentation \cite{LabelEfficient, DenoisingSegmentation} and object detection \cite{Diffusiondet}. Thanks to the great work DiffusionDet \cite{Diffusiondet}, our idea is largely inspired by the conclusion that relying on a denoising diffusion process from noisy boxes to object boxes, solving object detection in a generative way with diffusion models is effective and favorable. 

Notwithstanding the notable progress made, there has been no prior success in adapting generative diffusion models to visual grounding in the context of cross-modal perception. Such a gap is attributed to a compound challenge: how to adopt diffusion models to conduct cross-modal learning with the purposes of semantic alignment and localization? To the best of our knowledge, our paper on visual grounding is the first answer to this question.

\section{Methodology}
This section will first elaborate on the language-guided diffusion visual grounding process from the mathematical perspective, which insights will further be utilized as instruction for the design of our proposed methodology. 

\begin{figure*}[th!]
    \centering
    \includegraphics[width=\textwidth]{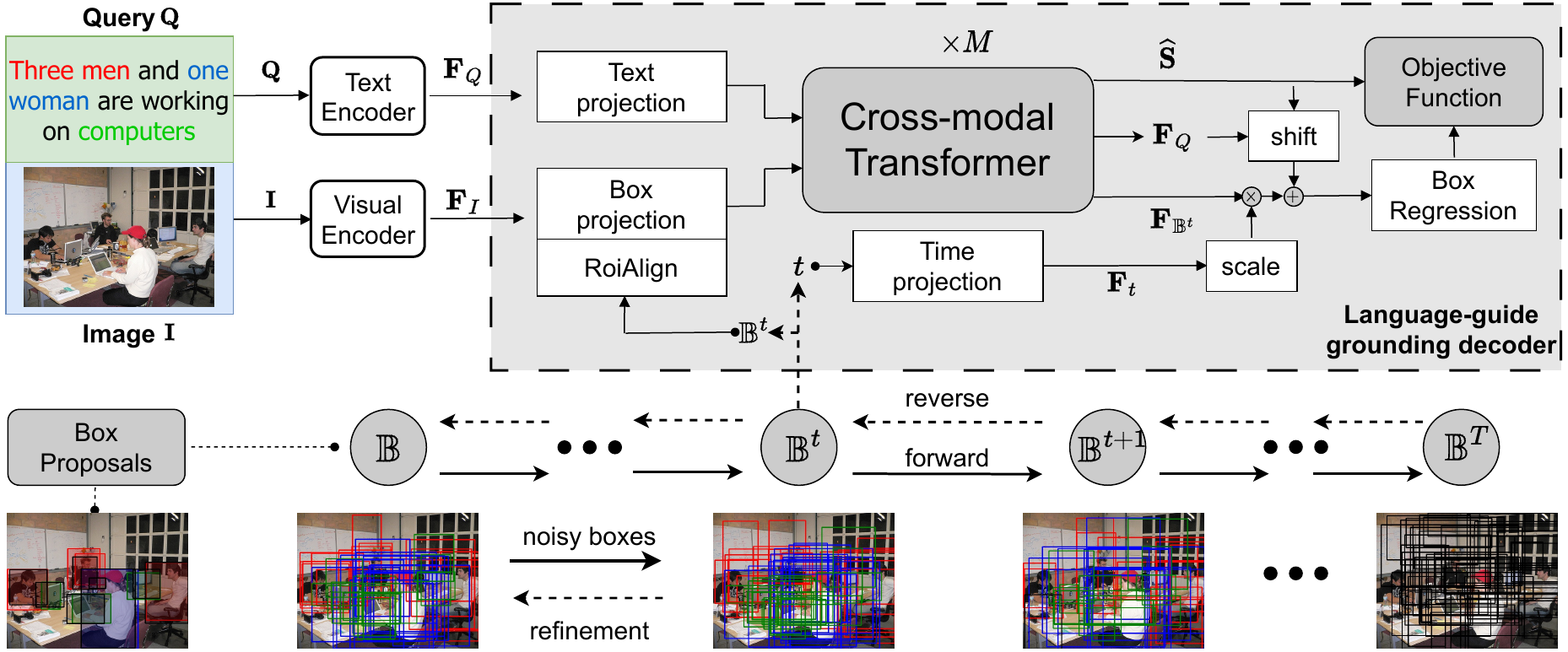}
    \caption{The language-guided noisy-to-box pipeline overview of our proposed LG-DVG. Given image $\bm{I}$ and the text query $\bm{Q}$ - "\textcolor{red}{Three men} and \textcolor{blue}{one woman} are working on \textcolor{red}{computers}", \textit{LG-DVG} begins by utilizing \textbf{visual and text encoders} to transform the input image and text queries into backbone features. Based on the vision backbone, features of \textit{noisy boxes} $\mathbb{B}^t$ at time step $t$ generated from a set of \textit{proposal boxes} can be directly extracted. Subsequently, the \textbf{novel cross-modal transformer} fuses box and text features to produce compact multi-modal representations. Eventually, conditioned on text representation, our \textbf{language-guided grounding decoder} optimized by a compound objective function performs box regression to progressively refine box predictions $\widehat{\mathbb{B}}^t$ from noisy boxes.}
    \label{fig:LGDVG}
    \vspace{-0.3cm}
\end{figure*}

\subsection{Problem Scenario}

Given raw image $\bm{I}\in R^{H \times W}$  and the text query $\bm{Q}=\left\{Q_i\right\}_{i=1}^P$ containing $P$ phrases, visual grounding aims to locate image regions $\mathbb{G} \subset \bm{I}$ for $\bm{Q}$ by a series of bounding boxes $ \mathbb{B}:\left\{\bm{B}_i\right\}_{i=1}^P$, where $\bm{B}_i=\left( {{\bm{b}_{i1}},{\bm{b}_{i2}},...,{\bm{b}_{iN_i}}} \right) \in {R^{N_i \times 4}}$ and $N_i \in Z_+$. The semantic content of each region enclosed by the bounding box ${\bm{b}_{in}}=\left(c_x^i, c_y^i, w^i, h^i\right)_{n \in N_i} \in \left[0, 1\right]^4$ that defines ground truth box center coordinates and its height and width relative to the image size, is the same as the phrase with index $i$. The one-to-many challenge occurs when $\exists N \in \left\{N_i\right\}_{i=1}^P$ satisfies $N > 1$. Our objective is to reformulate visual grounding by iterative learning, enabling small-scale randomly initialized boxes to be adjusted to approach ground truth regions progressively. 

\subsection{Mathematical Overview}

\noindent \textbf{Iterative paradigm defined by diffusion models}. By building our framework upon denoising diffusion models \cite{BasicDiffusion,DiffusionBeatGANs}, this paper naturally defines iterative learning for visual grounding as a generative task over the space of $\mathbb{B}$. These lead to our proposed \textit{language-guided diffusion model}. Without loss of generality, we omit the image $I$ in the derivation, since it is safe to assume that visual features are already included in $\bm{B}$.

Let $T$ denote the length of the Markov chain. Given ground truth boxes $\mathbb{B}^{t=0}$, which can be viewed as samples from a real data distribution, the \textbf{forward diffusion process} gradually adds Gaussian noise $\mathcal{N} \left(0, \bm{I}\right)$, governed by a variance schedule $\left\{\beta^t \in \left(0, 1\right)\right\}_{t=1}^T$ \cite{BasicDiffusion}, to bounding boxes in $T$ steps, producing a sequence of noisy boxes $\mathbb{B}^{t=1},...,\mathbb{B}^{t=T}$. As each $\bm{B}_i^{t} \in\mathbb{B}^t $ in step $t$ is the noisy sample from the specific $\bm{B}_i^{t=0}$, which corresponds to the $i$-th phrase $Q_i$, the forward noise process dominated by the text query should be defined as: 
\vspace{-0.1cm}
\begin{equation}
    \begin{aligned}
q\left(\bm{B}_i^{t}|\mathbb{B}^{0}, Q_i\right)=\mathcal{N}\left(\bm{B}_i^{t};\sqrt{\overline{\alpha}}\bm{B}_i^{0}, \left(1-\overline{\alpha}\right)\bm{I}\right)
    \end{aligned}
\label{eq:fdifusion}
\vspace{-0.1cm}
\end{equation}
where $\overline{\alpha}^t:=\prod_{j=0}^t \alpha^j$ and $\alpha^j = 1- \beta^j$.

Therefore, conditional on the phrase $Q_i$, the \textbf{reverse diffusion process}, presented as $\mathbb{B}^T \rightarrow...\rightarrow \mathbb{B}^t \rightarrow \mathbb{B}^{t-1} \rightarrow ... \rightarrow \mathbb{B}^0$, is to reconstruct the corresponding ground truth boxes $\bm{B}_i^0$ that are sampled from the conditional distribution $q\left(\bm{B}_i^0|Q_i\right)$ with the condition $Q_i$. To this end, each reverse step is to be defined as $q\left(\bm{B}_i^{t-1}|\mathbb{B}^t, Q_i\right)$. As this conditional probabilities cannot be estimated without using whole dataset, we introduce the approximation method proposed by work \cite{BasicDiffusion} to define the reverse diffusion process as 
\vspace{-0.1cm}
\begin{equation}
    \begin{aligned}
p_{\theta}\left(\mathbb{B}^{0:T}|Q_i\right)=p\left(\mathbb{B}^T\right)\prod_{t=1}^Tp_{\theta}\left(\bm{B}_i^{t-1}|\mathbb{B}^t, Q_i\right)
    \end{aligned}
\label{eq: reverse}
\vspace{-0.1cm}
\end{equation}
where $p_{\theta}$ with parameters $\theta$ is utilized to approximate conditional probabilities.

\noindent \textbf{Training}. After including the text query as input to guide the reverse process, we can directly incorporate conditions $\bm{Q}$ into a trainable model denoted as $f_{\theta}\left(\mathbb{B}^{t}, t, \bm{Q}\right)$. By carefully merging conditional information into the model learning as shown in work \cite{DiffusionBeatGANs}, the $f_{\theta}\left(\cdot\right)$ can be trained to predict $B_i^0$ from noisy boxes $B_i^t$ for $i=1,..,P$ \cite{BasicDiffusion}, thus leading to the objective function $\frac{1}{2}\sum_{i=1}^P ||f_{\theta}\left(\mathbb{B}^{t}, t, \bm{Q}_i\right) - B_i^0||^2$.

\noindent \textbf{Inference}. Following the iterative reconstruction rule proposed by work \cite{BasicDiffusion}, boxes for queries are to be produced by the reverse process shown in Eq. ~\ref{eq: reverse} in which proposal boxes $\mathbb{B}^T$ are generated from random Gaussian distributions. For efficiency concerns, we leverage DDIM \cite{DDIM} to perform inference.

\subsection{The Proposed Framework}

With the mathematical insights of the language-guided denoising diffusion process, LG-DVG naturally achieves iterative cross-modal reasoning but with a surprisingly lightweight architecture, as depicted in Figure~\ref{fig:LGDVG}.

\noindent \textbf{Proposal boxes}. In the inference stage, an arbitrary number of proposal boxes $\mathbb{B}^T$ are generated by sampling from Gaussian distribution $\mathcal{N} \left(0, \bm{I}\right)$. In the training stage, we generate a fixed set of $\widehat{N}$ proposal boxes by applying the over-sampling \cite{OverSampling} on ground truth boxes $\mathbb{B}:\left\{\bm{B}_i\right\}_{i=1}^P$. As an alternative to random sampling, we propose a phrase balanced sampling, which pads the box set by replicating boxes from a minority set $\bm{B}_k$, where $N_k =min\left\{N_1, ..., N_P\right\}$, to achieve $\widehat{N_k}=\frac{\widehat{N}}{P}$, $k \in [1, P]$. For simplify, we will still denote proposal boxes after padding as $\mathbb{B}:\left\{\bm{B}_i\right\}_{i=1}^P$ in which $|\bm{B}_i|=\widehat{N_i}$. Please refer to the appendix for detailed algorithm.

\noindent \textbf{Noisy boxes}. During training, the forward diffusion process, shown by Eq.~\ref{eq:fdifusion} of \textit{LG-DVG} involves adding Gaussian noise to proposal boxes $\mathbb{B}^{t=0}$ to compute noisy boxes $\mathbb{B}^{t}$ for a randomly selected time step $t$.

\noindent \textbf{Text encoder}. Given the text query $\bm{Q}$ containing $P$ phrases, the pretrained phraseBert \cite{PhraseBert} separately extracts features $\bm{F}_{Q_i} \in R^{d_t}$ for each phrase with index $i$, where $d_t = 768$. Alternatively, we obtain phrase features by averaging the final-layer token-level vectors yielded by BERT \cite{BERT} or Word2Vec \cite{W2V}. 

\noindent \textbf{Text projection}. The trainable text projection, which takes the form of Multilayer Perceptrons (MLPs), maps phrases feature $\bm{F}_{\bm{Q}} \in R^{P \times d_t}$ to a common space of dimension $D$, resulting in $\bm{F}_{\bm{Q}} \in R^{P \times D}$.

\noindent \textbf{Visual encoder}. The multi-scale feature maps $\bm{F}_I$ for the input raw image $\bm{I}$ are extracted by employing a pre-trained model, such as ResNet-related \cite{ResNet} and Swin transformer \cite{SwinTransformer} models, followed by a Feature Pyramid Network \cite{FeaturePyramid}. For efficiency, the visual encoder performs inference only once and keeps it frozen throughout the learning process.

\noindent \textbf{Boxes representation}. Boxes feature $\bm{F}_{\mathbb{B}} \in R^{\widehat{N} \times d_b}$ are obtained by cropping ROI-feature with RoIAlign \cite{RoIAlign} from the visual feature maps $\bm{F}_I$. After performing box projection, the obtained boxes representation is mapped to a common space of dimension $D$, resulting in $\bm{F}_{\mathbb{B}} \in R^{\widehat{N} \times D}$.

\noindent \textbf{Time projection}. For a time step expressed as a scalar, we utilize sinusoidal position embedding as in DiffusionDet to convert this discrete value into a continuous feature, which is then mapped by an MLP text projection to a space of dimension $D$, yielding $F_{t} \in R^{D}$. 

\begin{figure}[t]
    \centering
    \includegraphics[width=\columnwidth]{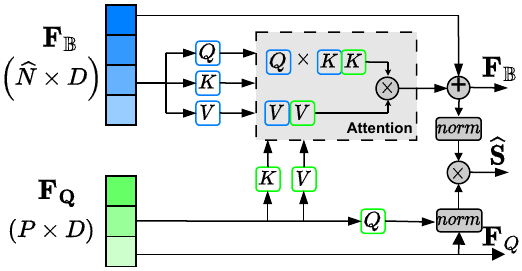}
    \caption{Our proposed cross-modal Transformer. We omit the layer norm and feedforward layer \cite{AllAttention} for brevity in this figure. }
    \label{fig:CrossTrans}
    \vspace{-0.1cm}
\end{figure}

\noindent \textbf{Cross-modal transformer}. Advanced diffusion models \cite{DALLE2, Imagen} for text-to-image generation aim to incorporate conditional text information into the model, thereby guiding image synthesis. However, effective language conditioning in the image-text alignment task requires merging text features into visual boxes for multi-modal fusion, accomplishing cross-modal relation learning, and maintaining the language information for further guidance. Thus, referring cross-attention module of Perceiver architecture \cite{Perceiver}, which projects the unchangeable input array to a latent bottleneck, our design of the language-guided grounding decoder is illustrated in Figure.~\ref{fig:CrossTrans}.

Specifically, after mapping $\mathbf{F}_{\mathbb{B}}$ and $\mathbf{F}_\mathbf{Q}$ to obtain the corresponding query, key and value terms, their keys, and values are concatenated to yield $\left[\mathbf{F}^k_{\mathbb{B}}, \mathbf{F}^k_{\mathbf{Q}}\right]$ and $\left[\mathbf{F}^v_{\mathbb{B}}, \mathbf{F}^v_{\mathbf{Q}}\right]$, respectively. These two new terms and $\mathbf{F}^q_{\mathbb{B}}$ are subsequently forward classical computation paradigms to produce the solid box features $\mathbf{F}_{\mathbb{B}}$. Eventually, as shown in Figure.~\ref{fig:CrossTrans}, to finalize cross-modal relations $\widehat{\mathbf{S}} \in R^{\widehat{N} \times P}$, box features and the text query $\mathbf{F}^q_\mathbf{Q}$ are multiplied, where norm computations derive from Segmenter \cite{Segmenter}. As in Perceiver \cite{Perceiver}, the language part $\mathbf{F}_\mathbf{Q}$ is maintained throughout this process. 

\noindent \textbf{Text conditioning box regression}. By incorporating the text query into the box regression, our text conditioning box regression makes predictions on noisy boxes to gradually approach phrase-align ground truth boxes with the input text reference. In order to accomplish this, we expand upon the shift-scale mechanism utilized in the class-conditional diffusion model \cite{DiffusionBeatGANs} and DiffusionDet \cite{Diffusiondet} by incorporating a text representation as a conditioning variable within the box regression. As shown in the box regression part of Figure.~22, the projection time $F_t$ is mapped to a \textit{scale} vector, while the output $\mathbf{F}_{\mathbf{Q}} \times \widehat{\mathbf{S}} \in R^{\widehat{N} \times D}$ of the cross-modal transformer is mapped to a \textit{shift} matrix, which behaves as the core condition information. Thus, box regression utilizes the $\mathbf{F}^v_{\mathbb{B}}\left(scale+1\right)+shift$ as input to predict $\left(\triangle c_x, \triangle c_y, \triangle w, \triangle h\right)$.

\subsection{Objective Function}
For a fixed-size set of $\widehat{N}$ noisy boxes $\mathbb{B}^{t}$ at time step $t$, \textit{LG-DVG} infers $\widehat{N}$ predictions $\widehat{\mathbb{B}}^{t=0}$, which is significantly larger than the number of ground truth boxes $N$ before over-sampling. Besides, as shown by Eq.~\ref{eq: reverse}, for the phrase query with index $i$, a subset of boxes $\widehat{\bm{B}}_i \subset \widehat{\mathbb{B}}^{t=0}$ is solely matched with ground truth boxes $\bm{B}_i$ to compute the loss function. To find bipartite matching between these two sets, we exploit the Hungarian matching algorithm discussed in work \cite{DETR}, thus having the objective function as: 
\begin{equation}
    \begin{aligned}
    \sigma_1, ..., \sigma_P = \underset{\sigma_1, ..., \sigma_P}{\arg\min}\sum_{i=1}^{P}\sum_{n=1}^{\widehat{N}_i}\mathcal{L}_{match}\left(\bm{b}_{in}, \bm{\widehat{b}}_{\sigma_i\left(n\right)}\right)
    \end{aligned}
\end{equation}
where $\sigma_i$ is a permutation of $\widehat{N}_i$ boxes. $\mathcal{L}_{match}$ is the matching cost function, such as Intersection over Union (IOU) score.

After computing the optimal one-to-one assignment $\sigma_1, ..., \sigma_P$ between predictions $\widehat{\mathbb{B}}^{t=0}$ and ground truth boxes $\mathbb{B}^{t=0}$, we define the objective function as a linear combination of a box loss and a cross-modal similarity auxiliary loss. 

Similarly to that of common grounding detectors \cite{TransVG}, the box loss contains smooth L1 loss and generalized IoU loss \cite{GIoU} between predictions and ground truth boxes. Thus, for a bounding box with index $j$ in $i$-th phrase, the loss is computed as $\mathcal{L}_{B_{in}} = \alpha \mathcal{L}_{smooth-l1}\left(\bm{b}_{in}, \bm{b}_{\sigma_i\left(n\right)}\right)+ \beta \mathcal{L}_{giou}\left(\bm{b}_{in}, \bm{b}_{\sigma_i\left(n\right)}\right)$, where $\alpha$ and $\beta$ are hyper-parameters.

The cross-modal similarity loss is an auxiliary term to force that semantic similarity between the text query and boxes is consistent with IoU scores between their bounding boxes. Thus, for a bounding box with index $j$ in noisy boxes $\mathbb{B}^t$, the loss is computed as $\mathcal{L}_{S_{j}} = \mathcal{L}_{l1}\left(\widehat{S}_{j},IoU\left(\mathbf{b}^t_{j}, \mathbb{B}\right)\right)$, where $\widehat{S}_{j} \in R^{P}$ is the output of cross-modal transformer, as illustrated in Figure.~\ref{fig:CrossTrans}.

Eventually, our objective function is defined as $\min \sum_{i=1}^P\sum_{n=1}^{\widehat{N}_i}\mathcal{L}_{B_{in}} + \lambda \sum_{j=1}^{\widehat{N}}\mathcal{L}_{S_{j}}$, where $\lambda$ is generally setup to be a small value to alance these two losses.

\section{Evaluation}

\noindent \textbf{Datasets}. The evaluation is performed on two distinct tasks: phrase localization, which encompasses the \texttt{Flickr30k Entities} \cite{Flickr30kE} dataset, and referring expression comprehension, which comprises \texttt{ReferItGame} \cite{RIG}, \texttt{RefCOCO} \cite{RefCOCO}, \texttt{RefCOCO+} \cite{RefCOCO}, and \texttt{RefCOCOg} \cite{RefCOCOg} datasets. 

\noindent \textbf{Inputs}. Consistent with the practical image augmentation techniques employed in prior studies \cite{TransVG, MDGT}, we resize the input image size to $640 \times 640$. Bert \cite{BERT} tokenizer is used to transform text samples, which are then padded with empty phrases up to the maximum phrase length among them. Due to space constraints, additional details are provided in the Appendix.

\noindent \textbf{Diffusion models}. We set the length of the Markov Chain to $1000$. The variance schedule $\beta$ in Eq.~\ref{eq:fdifusion} that governs the Gaussian noise scale added in each chain step is established as a monotonically decreasing cosine schedule \cite{ImprovedDenoising}. Additionally, the bounding boxes are normalized to the range of $[-scale, scale]$ in each step, where the signal scaling value $scale$ is set to $2.0$, following the discussion presented in \cite{Diffusiondet}.

\noindent \textbf{Training details}. We employ the pre-trained phraseBert \cite{PhraseBert} model from the Hugging Face \cite{HuggingFace} library. ResNet (RN) \cite{ResNet} and Swin transformer (SwinB) \cite{SwinTransformer} for the visual encoder are initialized with pre-trained weights on ImageNet-1K, and ImageNet-21K \cite{Imagenet}, respectively. These encoders are frozen without being trained. The resolution and sampling ratio of RoiAlign \cite{RoIAlign} are set to $7 \times 7$ and $2$, respectively. Our grounding decoder is initialized with Xavier init \cite{Xavier}. LG-DVG is trained with the AdamW optimizer with the initial learning rate as $10^{-4}$ and the weight decay as $10^{-4}$. Due to the simplicity of LG-DVG, the total number of trainable parameters, amounting to $7,805,444$, can be effectively trained for $60$ epochs within around $17$ hours using a mini-batch of $20$ on a single Tesla T4 GPU. We set the number of \textbf{Proposal boxes} to be $\widehat{N}=150$. The weights of losses are $\alpha = 2$, $\beta=5$, and $\lambda = 1$. We set $M=1$, which means that LG-DVG includes one cross-modal transformer.

\noindent \textbf{Inference details}. Box proposals with $\widehat{N}_{infer}=\left\{50, 100, 150, 200, 300, 800\right\}$ are generated by sampling from a Normal Gaussian distribution directly. Subsequently, a reverse diffusion process, demonstrated by Eq.~\ref{eq: reverse}, is carried out via DDIM \cite{DDIM} schema.

\noindent \textbf{Metrics details}. 
Our performance reporting adheres to the standard protocol \cite{MDGT, TransVG, VLTVG}, which involves presenting the top-1 accuracy (\%) metric. For a predicted bounding box to be considered correct, its intersection-over-union (IoU) with the ground-truth bounding box must exceed a threshold $\zeta \in \left[0.5, 0.95\right]$. When $\zeta=0.5$, the corresponding metric is $Acc@0.5$. In addition, we demonstrate the effectiveness of LG-DVG in handling one-to-many cases, where a single query may correspond to multiple ground-truth boxes, through classical qualitative results obtained from the \texttt{Flickr30k Entities} dataset \cite{Flickr30kE}. The inference time in milliseconds is also reported. The VGG network \cite{vgg}, Residual network \cite{ResNet}, SwimTransformer \cite{SwinTransformer} with the Swin-B structure, and Darknet are briefly abbreviated as VGG, RN, SwimB, and DN, respectively.

\subsection{Comparisons with State-of-the-art Methods}

Table \ref{tab:fk30} and Figure~\ref{fig:Qualitative} present the performance of LG-DVG on the \texttt{Flickr30k Entities} test set, a pretty challenging dataset for phrase localization because it involves one-to-many scenarios where each noun-phrase in a textual query to multiple ground truth boxes. By utilizing SwimB and PhraseBert (phr-bert) encoders, our LG-DVG model attains a top-1 accuracy of $79.92\%$, which is only marginally lower by $0.05\%$ compared to the state-of-the-art iterative approach M-DGT. M-DGT is a large-scale model relying on graph and transformer models to perform handcrafted reasoning, making it hard to be trained. However, built upon the Markov Chain of diffusion models, our LG-DVG with a lightweight structure with $7.8$M trainable parameters naturally implements progressive refinement step by step, as shown in Figure~\ref{fig:Qualitative}. Thus, the single-image inference time of the LG-DVG is $41.96ms$ lower than that of M-DGT. Compared to the TransVG, which represents the best one-stage approach, the LG-DVG achieves a significantly higher accuracy of $0.82$ while retaining a comparable inference time.

We report the performance of LG-DVG on referring expression comprehension task in Table \ref{tab:refer} and Figure~\ref{fig:Qualitative} containing results from \texttt{RefCOCO}/\texttt{RefCOCO+}/\texttt{RefCOCOg}/\texttt{RefCLRF}. It demonstrates that the LG-DVG outperforms alternative leading approaches on half splits of Refer-related datasets, and the average value of this excess is $0.14\%$. Significantly, LG-DVG is $2.04$ higher than the advanced iterative-based approach VLTVG on the TestB split of \texttt{RefCOCO+}. As illustrated by Figure~\ref{fig:Qualitative}, the underlying reason can be attributed to the fact that LG-DVG is mathematically designed to learn how to detect targets from noisy boxes, enabling it to generalize well across diverse object categories. Besides, on the \texttt{RefCELF} dataset containing more general natural scenes, LG-DVG maintains the best $72.82$ top-1 accuracy.

\begin{table}
\begin{center}
\caption{Comparison of phrase localization performance with state-of-the-art methods using the metric $Acc@0.5$ on the test set of \texttt{Flickr30k Entities} \cite{Flickr30kE}. The best performance is emphasized using bold black font. $\times$ means that the underlying method cannot use the assigned encoder.}
\label{tab:fk30}
\begin{adjustbox}{max width=\columnwidth}
\begin{tabular}{lcccc}
\toprule
Method & \begin{tabular}[c]{@{}c@{}}Visual \\ Encoder\end{tabular} & \begin{tabular}[c]{@{}c@{}}Text \\ Encoder\end{tabular} & Acc@0.5 & Time (ms)  \\

\hline
\multicolumn{5}{c}{\textit{Two-stage frameworks}}  \\ 
\hline

CCA \cite{Flickr30kE}                   &   VGG19           &  Word2vec     &  50.89    &  -     \\
SPC+PPC \cite{LanguageCues}             &   RN101       &  Word2vec     &  5.85     &  -       \\
SeqGROUND \cite{SeqGROUND}              &   RN50        &  LSTM         &  61.06    &  -       \\
CITE \cite{ConditionalEmbedding}        &   VGG16           &  Word2vec     &  61.89    &  184     \\
DDPN \cite{DDPN}                        &   RN101       &  LSTM         &  73.3     &  196      \\
LCMCG \cite{LCMCG}                      &   RN101       &  Bert         &  76.74    &  -       \\ 

\hline
\multicolumn{5}{c}{\textit{One-stage frameworks}}  \\ 
\hline

FAOA \cite{FAOA}                        &   DN53           &  Bert             &  68.69    &  38        \\
FAOA+RED \cite{DeconVG}                 &   DN53           &  Bert             &  70.50    &  -        \\
TransVG \cite{TransVG}                  &   RN50           &  Bert             &  78.47    &  -   \\ 
TransVG \cite{TransVG}                  &   RN101          &  Bert             &  79.10    &  61.77     \\
TransVG \cite{TransVG}                  &   RN101          &  Phr-Bert         &  $\times$        &  -     \\
\hline
\multicolumn{5}{c}{\textit{Iterative-based frameworks}}  \\ 
\hline

ReSC-large  \cite{RecursiveSubQuery}    &   DN53            &  Bert             &  69.28    &  36      \\
VLTVG \cite{VLTVG}                      &   RN50            &  Bert             &  79.18    &  -      \\
VLTVG \cite{VLTVG}                      &   RN101           &  Bert             &  79.84    &  -       \\
VLTVG \cite{VLTVG}                      &   RN101           &  Phr-Bert         &  $\times$    &  -       \\
M-DGT \cite{MDGT}                       &   RN50            &  Bert             &  79.32    &  91     \\
M-DGT \cite{MDGT}                       &   RN101           &  Bert             &  \textbf{79.97}    &  108     \\

\hline

LG-DVG                                  &   RN101           &  Bert          &  75.79    &  58.01           \\
LG-DVG                                  &   SwinB           &  Bert          &  76.62    &  66.76           \\
LG-DVG                                  &   RN50            &  Phr-Bert      &  79.03    &  45.1           \\
LG-DVG                                  &   RN101           &  Phr-Bert      &  79.21    &  57.42          \\
LG-DVG                                  &   SwinB           &  Phr-Bert      &  79.92    &  66.04           \\

\hline

\end{tabular}
\end{adjustbox}
\end{center}
\vspace{-0.9cm}
\end{table}

\begin{table*}[htb]
\begin{center}
\caption{Comparison with state-of-the-art methods on \texttt{RefCOCO} \cite{RefCOCO}, \texttt{RefCOCO+} \cite{RefCOCO}, \texttt{RefCOCOg} \cite{RefCOCOg}, and \texttt{ReferItGame} (\texttt{RefCLEF}) datasets in terms of metric $Acc@0.5$. Our LG-DVG is equipped with SwimB and Phr-Bert. The best performance is emphasized using bold black font.}
\label{tab:refer}
\begin{adjustbox}{max width=0.85\textwidth}

\begin{tabular}
{c@{\qquad}c@{\qquad}ccc@{\qquad}ccc@{\qquad}cc@{\qquad}c}
\toprule
\multirow{2}{*}{Method} & \multirow{2}{*}{Visual Encoder} & \multicolumn{3}{c}{ReferCOCO} & \multicolumn{3}{c}{ReferCOCO+} & \multicolumn{2}{c}{ReferCOCOg} & \multirow{2}{*}{RefCLEF } \\ \cline{3-9}
 &  & Val & TestA & TestB & Val & TestA & TestB & Val-u & Test-u \\ 

\hline
\multicolumn{10}{c}{\textit{Two-stage frameworks}}  \\ 
\hline

MAttNet \cite{MAttNet}  & RN101       &  76.65  &  81.14  &  69.99    &  65.33    &  71.62    &  56.02    &  66.58    &  67.27  & 29.04\\
A-ATT  \cite{AATT}      & VGG16       &  81.27  &  81.17  &  80.01    &  65.56    &  68.76    &  60.63    &  (73.18)  &  -      & -  \\ 
LGRANs \cite{LGRANs}    & VGG16       &  82     &  81.20  &  84.00    &  66.6     &  67.6     &  65.5     &  75.4     &  74.7   & - \\   
VS-graph \cite{VSGraph} & VGG16       &  82.68  &  82.06  &  84.24    &  67.70    &  69.34    &  65.74    &  75.73    &  75.31  & - \\ 
Ref-NMS \cite{RefNMS}   & RN101       &  80.70  &  84.00  &  76.04    &  68.25    &  73.68    &  59.42    &  76.68    &  76.73  & - \\ 
\hline 
\multicolumn{10}{c}{\textit{One-stage frameworks}}  \\ 
\hline   

FAOA \cite{FAOA}         & DK53       &  72.54  &  74.35  &  68.50    &  56.81    &  60.23    &  49.60    &  61.33  &  60.36   & 60.67 \\  
RCCF \cite{RealTimeCMC}  & DAL34      &  -      &  81.06  &  71.85    &  -        &  70.35    &  56.32    &  -      &  65.73   & 63.79 \\  
LBYL \cite{LBYLNet}      & DK53       &  79.67  &  82.91  &  74.15    &  68.64    &  73.38    &  56.94    &  62.70  &  -       & 67.47 \\   
TransVG \cite{TransVG}   & RN101      &  81.02  &  82.72  &  78.35    &  64.82    &  70.70    &  56.94    &  68.67  &  67.73   & 70.73\\   
LBYL+RED \cite{DeconVG}  & RN101      &  80.97  &  83.20  &  77.66    &  69.48    &  73.80    &  62.20    &  71.11  &  70.67   & 67.27 \\   
 
\hline
\multicolumn{10}{c}{\textit{Iterative-based frameworks}}  \\ 
\hline

Iterative \cite{IterativeSRL}        & RN101   &  -      &  74.27   &  68.10    &  -      &  71.05    &  58.25    &  -      &  70.05  & 65.48\\ 
ReSC-large \cite{RecursiveSubQuery}  & DK53    &  77.63  &  80.45   &  72.30    &  63.59  &  68.36    &  56.81    &  63.12  &  67.20  & 64.6\\ 
VLTVG \cite{VLTVG}                   & RN50    &  84.53  &  \textbf{87.69}   &  79.22    &  73.60  &  78.37    &  64.53    &  \textbf{74.90}  &  73.88  & 71.60\\ 
VLTVG \cite{VLTVG}                   & RN101   &  84.77  &  87.24   &  80.49    &  \textbf{74.19}  &  \textbf{78.93}    &  65.17    &  72.98  &  74.18  & 71.98\\ 

\hline

LG-DVG                         &   RN50   &  80.94  &  82.33  &  77.01    &  70.36   &  71.04    &  66.21 &  71.29  &  72.61   & 70.61\\
LG-DVG                         &   RN101  &  83.21  &  84.48  &  78.16    &  71.04   &  73.21    &  66.48 &  72.06  &  73.89   & 71.46\\
LG-DVG                       &   SwinB  &  \textbf{85.01}   &  86.63  &  \textbf{80.97}    &  73.17   &  75.82    &  \textbf{67.21} &  73.59  &  \textbf{75.31}   & \textbf{72.82}\\
\hline
\end{tabular}

\end{adjustbox} 
\end{center}

\vspace{-0.7cm}
\end{table*}

\begin{table}
\begin{center}
\caption{Comparison of the Top-1 accuracy (\%) metric under different $\zeta$ on the test set of \texttt{Flickr30k Entities} \cite{Flickr30kE} and \texttt{ReferItGame} \cite{RIG} datasets. Results obtained by our LG-DVG are highlighted with a gray background.}
\label{tab:tightness}
\begin{adjustbox}{max width=\columnwidth}
\begin{tabular}{ccccccc}

\toprule

Method & \begin{tabular}[c]{@{}c@{}}Visual \\ Encoder\end{tabular} & Acc@0.35 & Acc@0.5 & Acc@0.6 & Acc@0.7 &  Acc@0.9  \\
\hline

\multicolumn{7}{c}{\textit{\texttt{Flickr30k Entities}}}  \\ 
\hline

DDPN       & RN101                             & 83.72          & 73.3           & 64.71   & 50.81   & 30.36   \\
LCMCG      & RN101                             & 86.02          & 76.74          & 70.3    & 57.96   & 33.86   \\
FAOA       & DN53                              & 81.59          & 73.50          & 67.91   & 57.47   & 35.18   \\
TransVG    & RN101                             & 84.81          & 79.1           & 69.28   & 47.41   & 30.69   \\
M-DGT S=7  & RN101                             & \textbf{86.14} & \textbf{79.97} & 74.53   & 70.25   & \textbf{46.31}   \\
\rowcolor{gray!25} LG-DVG S=1 & SwimB          & 82.04          & 75.83          & 66.57   & 53.21   & 30.02   \\
\rowcolor{gray!25} LG-DVG S=3 & SwimB          & 83.93          & 79.21          & 71.42   & 70.06   & 40.04   \\
\rowcolor{gray!25} LG-DVG S=5 & SwimB          & 85.2           & 79.92          & \textbf{74.13}   & \textbf{70.15}   & 41.31   \\

\hline
\multicolumn{7}{c}{\textit{\texttt{ReferItGame}}}  \\ 
\hline

DDPN       & RN101          & 83.67    & 63.0    & 59.96   & 54.87   & 39.12   \\
FAOA       & DN53           & 86.93    & 60.67   & 53.76   & 46.21   & 31.68   \\
TransVG    & RN101          & 83.61    & 70.73   & 65.28   & 49.41   & 33.69   \\
ReSC-large & RN101          & 84.92    & 64.6    & 52.02   & 43.81   & 34.32   \\
\rowcolor{gray!25} LG-DVG S=1 & SwimB          & 81.78    & 69.81   & 56.13   & 42.26   & 34.24   \\
\rowcolor{gray!25} LG-DVG S=3 & SwimB          & 84.97    & 71.53   & 68.02   & 53.02   & 39.84   \\
\rowcolor{gray!25} LG-DVG S=5 & SwimB          & \textbf{86.1}     & \textbf{72.82}   & \textbf{68.95}   & \textbf{57.46}   & \textbf{42.82}   \\
\bottomrule
\end{tabular}
\end{adjustbox}
\end{center}
\vspace{-0.9cm}
\end{table}

\subsection{Main Properties}

\noindent \textbf{Simplicity}. First, with the mathematical guarantee from diffusion models with Markov Chain, the sole modules of LG-DVG to achieve iterative visual grounding are a cross-modal transformer and a shift-scale computation, whose inference costs around $2.08$ ms. Besides, a lightweight LG-DVG can be fully trained for $60$ epochs on a single GPU within 17 hours. Second, since LG-DVG models visual grounding as a generative task, box proposals employed during the inference stage are generated via random sampling from a Gaussian distribution.

\begin{figure}[t]
    \centering
    \includegraphics[width=0.9\columnwidth]{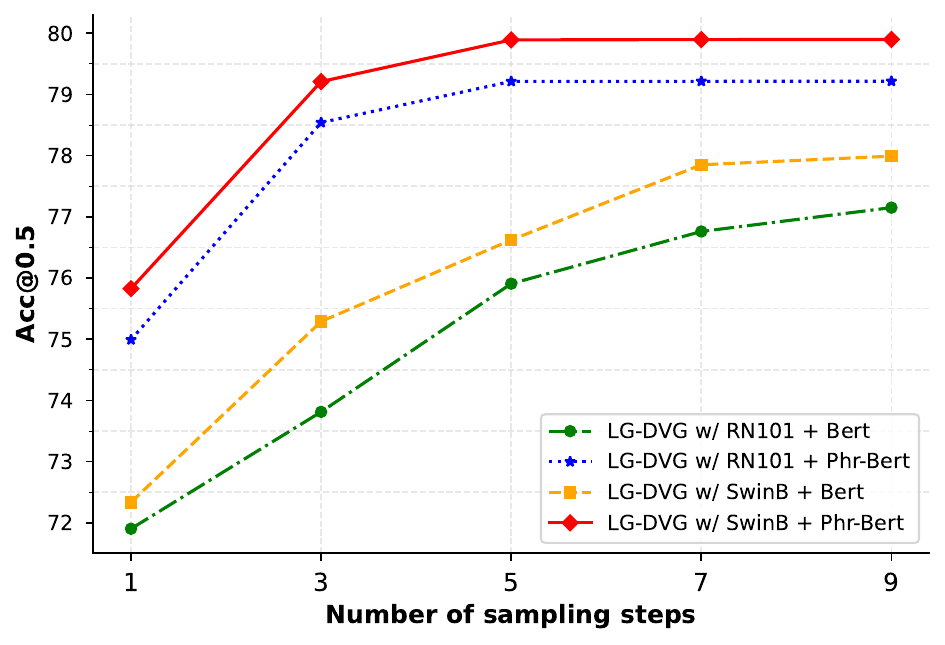}
    \caption{Comparison of the progressive refinement property of LG-DVG on \texttt{Flickr30k Entities} \cite{Flickr30kE} dataset using different visual and text encoders. For all cases, the accuracy increases with refinement times.}
    \label{fig:SamplingStep}
    \vspace{-0.5cm}
\end{figure}

\begin{figure}[t]
    \centering
    \includegraphics[width=\columnwidth]{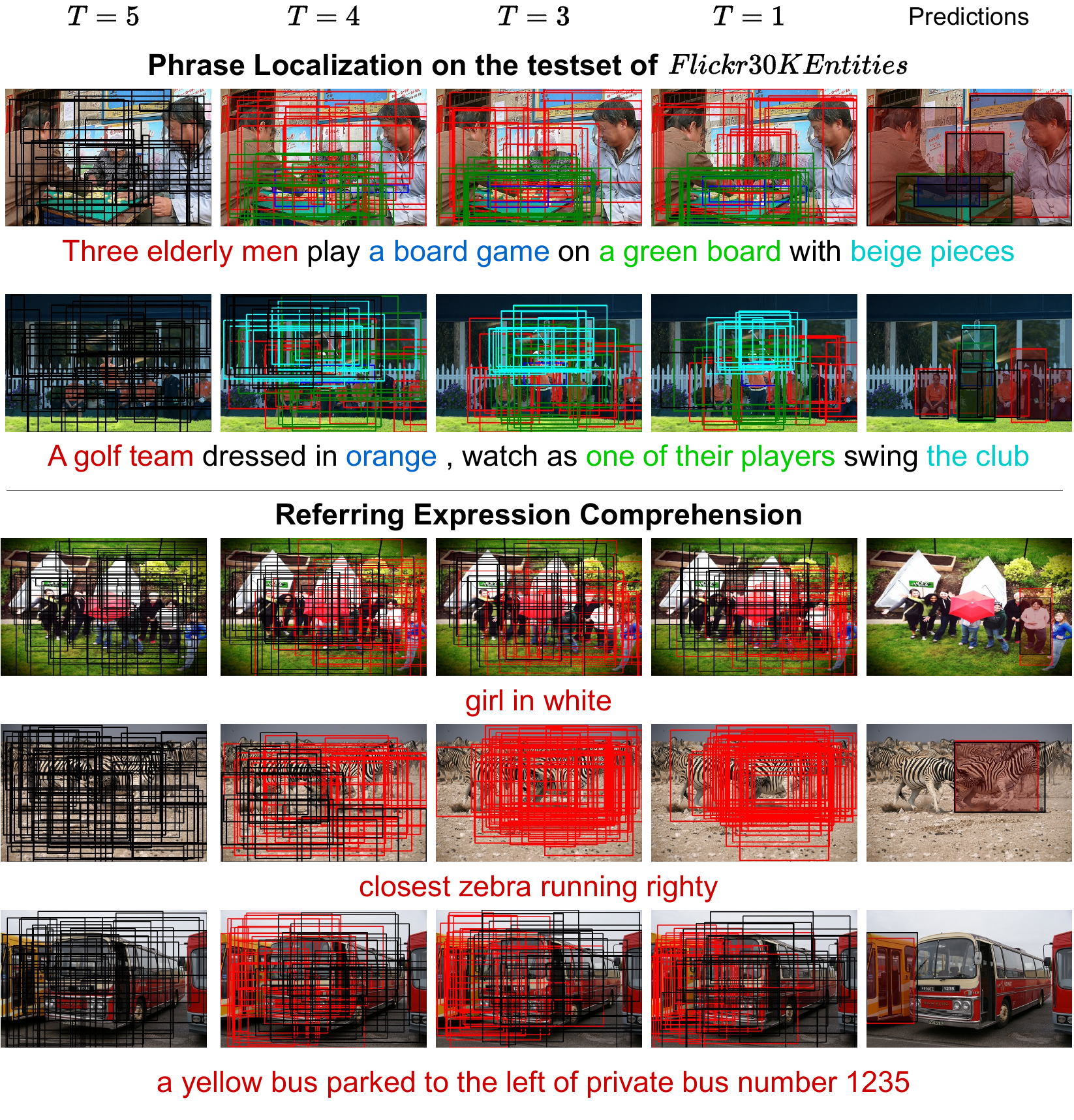}
    \caption{Qualitative results illustrating how LG-DVG progressively refines noisy boxes ($T=5$) conditional on the language query to approach ground truth boxes ($T=1$). In the fifth column, we present the ground truth boxes as the black rectangle to compare with our predictions. The results presented in the third, fourth, and fifth rows derive from test sets of \texttt{RefCOCO}/\textit{google}, \texttt{RefCOCO+}/\textit{unc}, and \texttt{refcocog}/\textit{umd}, respectively. For ease of presentation, we sample top $50$ boxes with the largest $\widehat{S}$ from $150$ proposal boxes.}
    \label{fig:Qualitative}
    \vspace{-0.1cm}
\end{figure}

\noindent \textbf{Progressive refinement}. The results presented in Figure~\ref{fig:SamplingStep}, the shaded part of Table~\ref{tab:tightness}, and Figure~\ref{fig:Qualitative}, indicate that the performance of LG-DVG steadily improves as the number of sampling steps increases from $1$ to $9$. Specifically, the average Top1 accuracy of LG-DVG's variations increases from $73.76425$ to $78.5645$. Besides, as shown by results in Table~\ref{tab:tightness} and Figure~\ref{fig:Qualitative}, LG-DVG with more sampling steps tends to produce tighter bounding boxes with refinement. For example, the $Acc@0.7$ of LG-DVG with more than $3$ sampling steps maintain $70.15$ and $57.46$ in \texttt{Flickr30k Entities} \cite{Flickr30kE} and \texttt{ReferItGame} \cite{RIG} datasets, respectively. 

\noindent \textbf{Tightness bounding boxes}. Compared with other advanced approaches, the results in Table~\ref{tab:tightness} demonstrate that the Top1 accuracy of iterative-based frameworks, such as M-DGT \cite{MDGT} and LG-DVG, decreases slower while maintaining a lower magnitude. Particularly, the Acc@0.5 of LG-DVG dropped by $9.77$ compared with Acc@0.5 in FK30 dataset, and the corresponding figure obtained in the REF dataset is 15.36. Nevertheless, when evaluated on the \texttt{Flickr30k Entities} \cite{Flickr30kE} and \texttt{ReferItGame} \cite{RIG} datasets, the leading one-stage approach, TransVG \cite{TransVG}, experiences a notable decline in performance metrics, with scores decreasing from $79.1$ to $47.41$ and $70.73$ to $49.41$, respectively. 

\subsection{Ablation Study}

\noindent \textbf{Sampling steps}. Figure ~\ref{fig:SamplingStep} demonstrates that for LG-DVG variations, increasing the DDIP sampling steps in the reverse step for inference leads to direct improvement on Acc@0.5. All variations generally reach the optimal within $9$ sampling steps. 

\noindent \textbf{Encoders}. The text encoder determines the performance in terms of accuracy and prediction speed of our LG-DVG, while the visual encoder has a lower influence. For example, Table~\ref{tab:fk30} presents that compared to the $0.2$ increase by changing from RN50 to RN101, converting from Bert to Phr-Bert direct gain $3.3$ and $3.42$ improvement for LG-DVG with SwimB and RN101, respectively. Figure~\ref{fig:SamplingStep} highlights that LG-DVG variations with Phr-Bert encoder achieve an accuracy of $79.568$ on average within $5$ sampling steps. Others with Bert perform more steps but gain a lower $77.572$ accuracy on average. 

\noindent \textbf{Components}. The significance of the language-guide grounding decoder (LG) to our LG-DVG is highlighted by replacing it with a combination of normal transformers and a direct box regression, such as the structure of FAOA \cite{FAOA}. In the absence of LG, the new DVG architecture achieves an Acc@0.5 performance lower than $70$ and experiences a drastic decline in performance at Acc@0.7. These results and the importance of Phr-Bert corroborate each other to illustrate the necessity of language-guided design of our LG-DVG. Furthermore, when incorporating DDIM, LG-DVG is able to achieve higher accuracy with a reduction of $4$ sampling steps.

\noindent \textbf{Boxes proposal schemas}. The outcomes reported in Table  \ref{tab:ablation} show that in the training stage, using phrase balanced sampling schema to pad ground truth boxes to obtain BBB with the number of $\widehat{N}$ is favorable for achieving optimal accuracy with minimum DDIM sampling steps. Padding with randomly generated boxes can include noisy boxes as targets for learning, thus damaging LG-DVG. Random over-sampling schema \cite{OverSampling} exhibits a notable of $6.818$, attributed to the fact that phrase query with majority ground truth boxes will dominate the model optimization.

\noindent \textbf{$\widehat{N}_{infer}$}. Increasing the number of proposal boxes $\widehat{N}_{infer}$ for inference yields improved accuracy results and tight bounding boxes for LG-DVG. However, these improvements come at the expense of increased inference time. Our setup N=150 achieves a better balance between inference time and accuracy. One interesting observation is that the chosen $\widehat{N}_{infer} \geq 200$ will contribute more to producing tight bounding boxes, leading to a larger gain in Acc@0.7.

\begin{table}
\begin{center}
\caption{Ablation studies of LG-DVG in \texttt{Flickr30k Entities} \cite{Flickr30kE} dataset. The best setup is emphasized using the gray background.}
\label{tab:ablation}
\begin{adjustbox}{max width=\columnwidth}
\begin{tabular}{c|c|c|c|c|c}
\hline
\multicolumn{6}{|c|}{Schemas for boxes proposal during training}                                    \\ 
\hline
\multicolumn{4}{c|}{Methods}                        & Acc@0.5          & Sampling steps                      \\ 
\hline
\multicolumn{4}{l|}{random generation}              & 62.961           & 12                         \\ 

\multicolumn{4}{l|}{random over-sampling}           & 73.106           & 6    \\ 

\rowcolor{gray!25} \multicolumn{4}{l|}{phrase balanced over-sampling}       & 79.924           & 5     \\ 
\hline
\hline
\multicolumn{6}{|c|}{Ablation of components} \\ 

\hline

    DDIM                   & \multicolumn{2}{c|}{LG}          & Acc@0.5          & Acc@0.7      & Sampling Steps    \\ 
\hline

                           & \multicolumn{2}{c|}{}              & 69.17             & 53.201      & 3            \\ 
                           & \multicolumn{2}{c|}{\checkmark}    & 77.983            & 64.383      & 9                \\
\checkmark                 & \multicolumn{2}{c|}{}              & 68.861            & 51.463      & 3                  \\
\rowcolor{gray!25} \checkmark                 & \multicolumn{2}{c|}{\checkmark}    & 79.924            & 70.152      & 5                  \\                           
\hline
\hline
\multicolumn{6}{|c|}{Proposal Boxes Count $\widehat{N}_{infer}$ during Inference} \\ 

\hline
\multicolumn{3}{c|}{$\widehat{N}_{infer}$}      & Acc@0.5      & Acc@0.7              & Time (ms)    \\ 
\hline
\multicolumn{3}{c|}{50}                          & 73.259       & 64.12                & 57.24      \\
\multicolumn{3}{c|}{100}                         & 78.647       & 70.011               & 59.89      \\
\rowcolor{gray!25} \multicolumn{3}{ c|}{150}                         & 79.924       & 70.152               & 66.04       \\ 
\multicolumn{3}{c|}{200}                         & 79.927       & 71.051               & 71.21       \\
\multicolumn{3}{c|}{300}                         & 79.93        & 71.385               & 87.57        \\ 
\multicolumn{3}{c|}{800}                         & 79.939       & 71.873               & 238.93    \\      
\hline
\end{tabular}
\end{adjustbox}
\end{center}

\vspace{-1cm}
\end{table}

\section{Conclusion}
In this paper, we present LG-DVG, a novel language-guided diffusion model that builds cross-modal reasoning with the Markov Chain to achieve iterative visual grounding. Through formulating region-text alignment as a denoising diffusion process from noisy boxes to target boxes conditional on the text query, LG-DVG is trained to reverse this process, thus gaining the property of gradually refining noisy boxes to approach query-related object boxes. This contributes to the lightweight, capable of attaining tight bounding boxes, and easy-to-use benefits of LG-DVG. Extensive experiments indicate that our LG-DVG shows favorable potential in terms of accuracy and speed compared to well-established frameworks. An attractive contribution brought by our model is to verify the effectiveness of addressing cross-modal alignment in a generative way.

{\small
\bibliographystyle{ieee_fullname}
\bibliography{main}
}

\clearpage
\appendix

\section{Visual Grounding via Iterative Reasoning}

The purpose of visual grounding is to establish correspondence between a given text query and particular regions within an image, subject to the condition that the resulting alignment of text and region contains consistent semantics. Nonetheless, searching for an aligned text-region pair can be challenging, as the image can contain an infinite number of regions. Despite not being classified as an NP-hard problem, cross-modal alignment needs to be decomposed and downgraded to achieve a computationally feasible solution. 

Iterative visual grounding, a novel direction emphasized by advanced methods \cite{VLTVG, MDGT}, introduces iterative reasoning, which converts a complicated cross-modal alignment problem into many easy-to-solve sub-problems that can be addressed progressively to obtain robust results with high performance. With such a heuristic learning process, no well-prepared box proposals from off-the-shelf detectors or dense anchors of various sizes and scales are needed. Instead, starting from noisy or imprecise boxes, the grounding method can gradually refine the output of each iteration to facilitate subsequent runs, thus eventually significantly improving the accuracy.

Typically, prior methods with once-for-all reasoning, including the two-stage and one-stage frameworks, can be converted to iterative visual grounding by running the learning process multiple times. For instance, an iterative version of FAOA \cite{FAOA} can utilize the box predictions from the previous iteration as the input anchors to be merged with the text query to make box repression. Similarly, such iterative learning architectures can be utilized for LCMCG \cite{LCMCG} approach. However, frameworks, especially TransVG \cite{TransVG}, fail to have an iterative version because their learning procedure is specifically designed with sequence learning to perform the once final reasoning. However, we argue that these iterative versions of approaches with once-for-all reasoning may be unlikely to improve grounding accuracy with the increase in iteration steps, as shown in Tabel~\ref{tab:fk30sota} and Figure~\ref{fig:morecompare}. 

\section{Highlights of LG-DVG}

Although alternative methods, typically VLTVG \cite{VLTVG} and M-DGT \cite{MDGT}, have attained top-tier performance through the adoption of iterative visual grounding, their structures are intricate and large-scale in terms of both size and training paradigm. Particularly, as the design of the iterative process heavily relies on human priors, these approaches are trapped by the issues of hard-to-train, long inference time, and lower generality. However, our proposed language-guided diffusion model, referred to as LG-DVG, is naturally built upon the Markov Chain to reformulate the cross-modal alignment into a query-guided noisy-to-box process with a rigorous theoretical guarantee from diffusion models \cite{BasicDiffusion, DiffusionBeatGANs}. This introduces multiple highlights.

\textbf{Lightweight model}. The iterative learning process of LG-DVG, which eliminates the need for human priors or manually designed components, is straightforward as one single grounding decoder solely conducts cross-modal learning and box regression step-by-step along the Markov Chain. Thus, with a mere 7.8 million trainable parameters, the entire model could be trained rapidly on a single GPU, requiring only 17 hours. The lightweight model and learning process employed in LG-DVG are conducive to achieving convergence, in contrast to the arduous training issue mentioned by the M-DGT \cite{MDGT}.

\textbf{Efficiency}. The use of iterative approaches incurs a comparatively higher inference time cost, exemplified by the $108$ milliseconds required by M-DGT \cite{MDGT}, which is even similar to that of two-stage frameworks. Even though VLTVG \cite{VLTVG} does not report inference time, the inference time can be even larger because it includes multi-stage reasoning with multiple complex modules. However, our LG-DVG with optimal setup requires only $66.04$ milliseconds, thereby establishing a new benchmark for iterative visual grounding.

\begin{figure*}[ht]
    \centering
    \includegraphics[width=\textwidth]{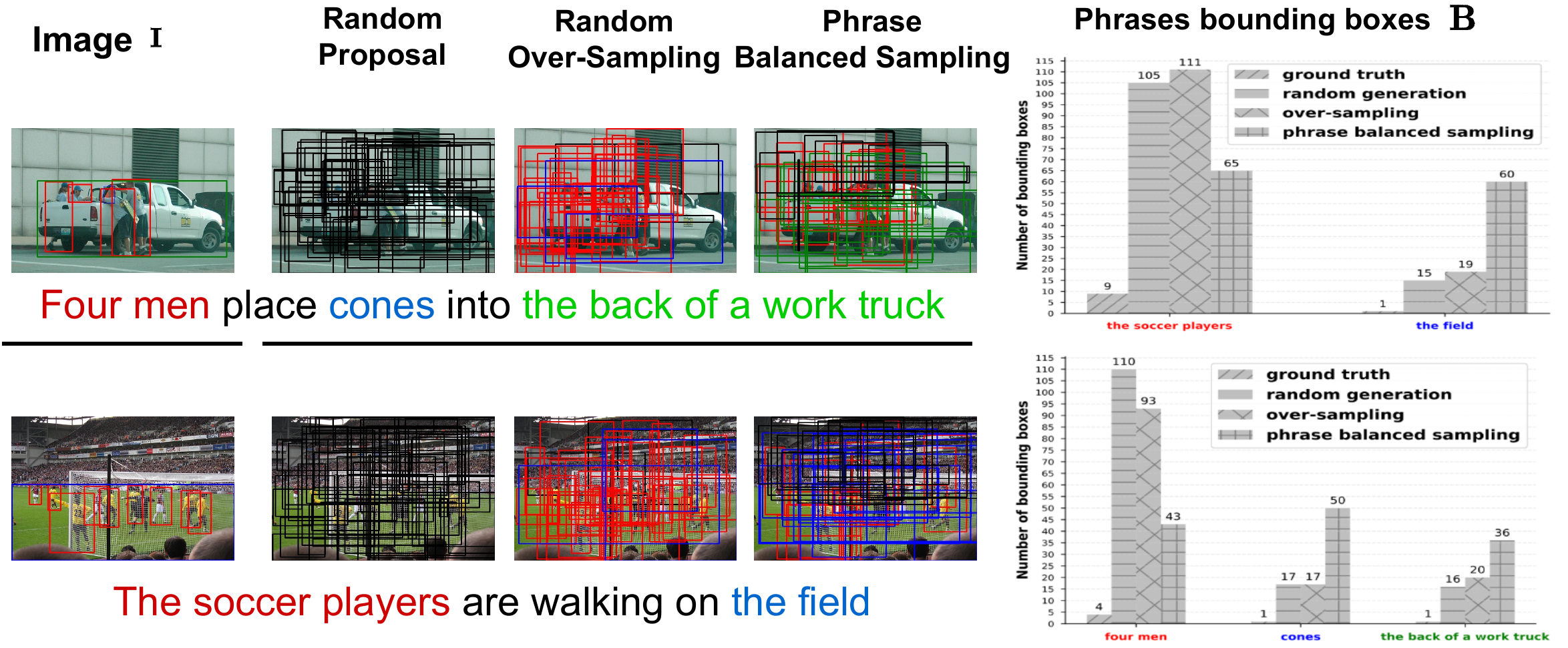}
    \caption{The illustration of how different box proposal mechanisms work.}
    \label{fig:phraseSampling}
    \vspace{-0.1cm}
\end{figure*}

\textbf{Tight bounding boxes}. Due to the progressive refinement property inherent in LG-DVG, Intersection over Union (IOU) scores between the predicted bounding boxes and ground truth boxes are relatively high. In contrast to the convoluted mechanism of M-DGT \cite{MDGT}, LG-DVG employs a conditional reasoning process within the Markov Chain, wherein box refinement in each step is guided by the language query and past predictions, thus achieving this outcome more naturally. The foremost competing methodology, VLTVG \cite{VLTVG}, does not report such an advantage. We demonstrate this benefit of LG-DVG in Section 8.

\textbf{One-to-many challenge}. When one query corresponds to multiple regions in the images, such as samples in the \texttt{Flickr30K Entities} dataset, our LG-DVG parallel predicts bounding boxes to align with each ground truth box of this query, which successfully addresses the one-to-many challenge. However, VLTVG \cite{VLTVG}, which produces only one box prediction for each image, fails to locate all potential regions for the text query. Such a great advantage of our approach can be demonstrated by presenting results in which, for a text query, the predicted bounding boxes locate at least $50\%$ of ground truth regions, and the IOU score between each prediction and ground truth one is higher than a threshold value $\zeta=0.5$. Ample evidence can be observed in Figures~\ref{fig:onetofive}, \ref{fig:onetoseven}, \ref{fig:onetonine}, \ref{fig:onetoeleven}, \ref{fig:oneto13}, and \ref{fig:oneto15}, which are obtained through the inference of the LG-DVG method on the test samples from the \texttt{Flickr30k Entities} data \cite{Flickr30kE}.

\section{Phrase Balanced Box Proposal}
During the training stage, the ground truth boxes $\mathbb{B}$ of phrases need to be padded to a predetermined size $\widehat{N}$. Subsequently, the forward diffusion process can gradually transform them into noisy boxes by adding Gaussian noise. There are mainly three mechanisms capable of sampling from $\mathbb{B}$ to achieve such a box proposal. 

\noindent \textbf{Random generation}. The most straightforward way is to randomly generate bounding boxes to be added to the ground truth $\mathbb{B}$ until the size reaches $\widehat{N}$. Although randomly generated boxes may not be correlated with any text query, the predictions of LG-DVG for these boxes are employed to calculate the loss with the ground truth, thereby ultimately contributing to the learning process. The DiffusionDet \cite{Diffusiondet} utilizes such a way to implement the box proposal and achieves the best performance. Conversely, this sampling scheme is unsuitable for our proposed LG-DVG. As shown by the random generation column of Figure~\ref{fig:phraseSampling}, dense bounding boxes are generated, and most do not align with ground truth boxes.

\noindent \textbf{Random over-sampling}. Through repeatedly duplicating the random boxes within the ground truth ones, random oversampling pads the $\mathbb{B}$ with its subsets until the size reaches $\widehat{N}$. In unbalanced machine learning, over-sampling \cite{OverSampling} is a general way to increase the size of samples. As evident from the random oversampling column in Figure \ref{fig:phraseSampling}, a group of bounding boxes is presented in which distinct subsets correspond to diverse phrases. However, such a sampling schema will exacerbate the bounding box imbalance among phrase queries. For instance, the text query for the image presented in the first row of Figure~\ref{fig:phraseSampling} contains three phrases corresponding to 4, 1, and 1 ground truth boxes, respectively. After random over-sampling for the box proposal, the $110$ bounding box of the first phrase query will dominate the learning process. The second row of Figure~\ref{fig:phraseSampling} depicts an image comprising 9:1 bounding boxes for two distinct phrases, which poses an even more significant challenge.

\noindent \textbf{Phrase balanced sampling}. We can achieve phrase balance sampling by considering the unbalanced bounding boxes of diverse phrases when applying oversampling mechanism. To elaborate, for an image containing $P$ phrase queries, each associated with a bounding box set $\mathbf{B}$, the phrase with the smallest $\mathbf{B}$ size is assigned the highest priority. Consequently, oversampling is conducted on this phrase until the size of its box proposals reaches the value of $\frac{\widehat{N}}{P}$. After performing multiple rounds of such sampling, the entire $\mathbb{B}$ will have a size of $\widehat{N}$, resulting in equivalent quantities of ground truth boxes corresponding to different phrases. As shown in Figure~\ref{fig:phraseSampling}, phrases presented in the first row eventually have $43$, $50$, and $36$ bounding boxes, respectively. The qualitative outcomes depicted in the final column corroborate the efficacy of phrase-balanced sampling in generating a uniform quantity of target boxes for diverse phrases.

\section{Learning Procedure}

In the training phrase, with the image and text as input, the LG-DVG builds box proposals before performing the diffusion process to transform them into noisy boxes. The model is trained to recover box proposals in the language-guided reverse diffusion process. Algorithm~\ref{algo:lgdvg-train} provides the pseudo-code of the training procedure. 

In the inference phrase, LG-DVG first generates $N_{infer}$ box proposals by sampling from a normal distribution $\mathcal{N}\sim \left(0, 1\right)$. Given the image and text, the languaged-guided denoising sampling process converts noisy boxes to ground truth boxes by progressively refining the predictions along the Markov Chain, as shown in Algorithm~\ref{algo:lgdvg-inference}. Specifically, for convenience, we denote reverse steps in the inference stage as $T$ because this is the sampling length of the Markov Chain. Thus, $T=4$ is the next reverse sampling step of $T=5$, and $T=1$ means the end of reverse sampling.

\begin{algorithm}[ht]
\SetKwInOut{Input}{input}
\SetKwInOut{Output}{output}
\SetAlgoLined

    \Input{Image $\bm{I}$ and the text query $\bm{Q}=\left\{Q_i\right\}_{i=1}^P$. The ground truth boxes $\mathbb{B}: \left\{\bm{B}_i\right\}_{i=1}^P$. Chain Steps $T=1000$. \#proposal boxes $\widehat{N}$. $\alpha$, $\beta$, and $\lambda$.}
    \KwOut{The computed loss $\mathcal{L}$.}

    \PyComment{Prepare box proposals via `PhraseBalancedSampling`} \\
    $\left\{\bm{B}^0_i\right\}_{i=1}^P$ = PBSampling($\left\{\bm{B}_i\right\}_{i=1}^P$, $\widehat{N}$)\\
    \PyComment{Visual and Text encoding} \\
    pretrained = True \\
    $F_I$ = image\_encoder($\bm{I}$, pretrained)\\
    \PyComment{$F_Q\in R^{P\times d_t}$}  \\
    $F_Q$ = text\_encoder($\bm{Q}$, pretrained)\\
    \PyComment{Forward diffusion process}\\
    \PyComment{1. generate time steps}\\
    $t$ = randint($0$, $T$)\\
    \PyComment{2. generate noise}\\
    noises = normal(mean=0, std=1) \\
    \PyComment{3. Languaged-Guided Forward Diffusion Process}\\
    $\mathbb{B}_n^t: \left\{\bm{B}^{t}_i\right\}_{i=1}^P$ = forward($\left\{\bm{B}^0_i\right\}_{i=1}^P$, $t$, noises)\\
    \PyComment{Text Projection $F_Q\in R^{P\times D}$}\\
    $F_Q$ = text\_proj($F_Q$)\\
    \PyComment{Perform Prediction}\\
    \PyComment{via language-guided grounding decoder}\\
    $\widehat{\mathbb{B}}^{t=0}$, $\bm{\widehat{S}}$ = lg\_decoder($\left\{\bm{B}^{t}_i\right\}_{i=1}^P$,$F_I$,$F_Q$,$t$) \\
    \PyComment{Hungarian matching}\\
    $\sigma_1, ..., \sigma_P$ = match($\left\{\widehat{\bm{B}}_i\right\}_{i=1}^P$, $\left\{\bm{B}_i\right\}_{i=1}^P$) \\
    \PyComment{Box loss}\\
    $L_{\bm{\widehat{B}}}$ = 0\\
    \PyCode{for $i$ in range($P$):} \\
    \Indp  
    $\mathcal{L}_{\bm{\widehat{B}}}$+=box\_loss($\sigma_i$, $\widehat{\bm{B}}_i$, $\bm{B}_i$, $\alpha$, $\beta$)\\
    \Indm 
    \PyComment{Similarity loss $\mathcal{L}_{\bm{\widehat{S}}}$}\\
    $\bm{\nu}$=zeros$\left(\widehat{N}, P\right)$\\
    \PyCode{for $n$ in range($\widehat{N}$):} \\
    \Indp  
        $\bm{b}_n = \mathbb{B}_n^t$\\
        \PyCode{for $i$ in range($P$):} \\
        \Indp 
        \PyCode{$\bm{\nu}[n, i]= \min IoU\left(\bm{b}_n, \bm{B}_i\right)$}\\
        \Indm
    \Indm 
    \PyCode{$\mathcal{L}_{\bm{\widehat{S}}} = l_1\left(\bm{\widehat{S}}, \bm{\nu}\right)$}\\
    \PyComment{Global loss $L$}\\
    $\mathcal{L} = \mathcal{L}_{\bm{\widehat{B}}}+ \lambda \mathcal{L}_{\bm{\widehat{S}}}$\\
    return $\mathcal{L}$
\caption{Training stage of LG-DVG}
\label{algo:lgdvg-train}
\end{algorithm}

\begin{algorithm}[ht]
\SetKwInOut{Input}{input}
\SetKwInOut{Output}{output}
\SetAlgoLined

    \Input{Image $\bm{I}$ and the text query $\bm{Q}=\left\{Q_i\right\}_{i=1}^P$. Chain Steps $1000$. \#sampling steps $n\_steps$, $is\_ensemble$.}
    \KwOut{The predicted boxes $\widehat{\mathbb{B}}^T$} 

    \PyComment{Visual and Text encoding} \\
    pretrained = True \\
    $F_I$ = image\_encoder($\bm{I}$, pretrained)\\
    \PyComment{$F_Q\in R^{P\times d_t}$}  \\
    $F_Q$ = text\_encoder($\bm{Q}$, pretrained)\\
    \PyComment{Text Projection $F_Q\in R^{P\times D}$}\\
    $F_Q$ = text\_proj($F_Q$)\\
    \PyComment{Uniform sample step size} \\
    times = reversed(linespace($-1$, $1000$, $n\_steps$)) \\
    \PyComment{$\left[(T-1, T-2), (T-2, T-3), ..., (0, -1)\right]$} \\
    time\_pairs = list(zip(times$\left[:-1\right]$, times$\left[1:\right]$) \\
    \PyComment{Prepare box proposals} \\
    $boxes$ = normal(mean=0, std=1) \\
    \PyComment{Prepare ensemble predictions} \\
    $ensem\_boxes$ = list() \\
    $ensem\_scores$ = list() \\
    \PyCode{for $cur\_t$, $next\_t$ in zip(time\_pairs):} \\
    \Indp 
        \PyComment{Predict $\mathbb{B}^0$ from noisy boxes} \\
        $boxes\_pred$, $\bm{\widehat{S}}$ = lg\_decoder($boxes$,$F_I$,$F_Q$,$t$) \\
        \PyComment{Estimate $boxes$ at $next\_t$} \\
        $boxes$ = ddim\_step($boxes$, $boxes_pred$, $cur\_t$, $next\_t$)\\
        if $is\_ensemble$: \\
        \Indp 
            $ensem\_boxes$.append($boxes\_pred$) \\
            $ensem\_scores$.append($\bm{\widehat{S}}$) \\
        \Indm  
    \Indm 
    if $is\_ensemble$: \\
    \Indp
        return $ensem\_boxes$, $ensem\_scores$\\
    \Indm  
    return [$boxes$], [$\bm{\widehat{S}}$]    
    
\caption{Inference stage of LG-DVG}
\label{algo:lgdvg-inference}
\end{algorithm}

\section{Limitations}
We argue that a significant constraint of the LG-DVG model lies in its inadequacy to effectively incorporate the semantic associations embedded within textual queries, which could potentially enhance the cross-modal reasoning capabilities. For instance, in the \texttt{Flickr30K Entities} dataset, given the text query "\textit{A group of businessmen walking down the road}" with two noun phrases, "\textit{A group of businessmen}" and "\textit{the road}", when two phrases exhibit a robust connection, exemplified by "walking down," incorporating such conjunctions can embrace a dual reasoning benefit, which allows the localization of phrases to mutually reinforce one another, thereby enhancing the overall accuracy and efficiency. However, LG-DVG omits this strong connection because it performs reasoning conditional on each phrase separately. Similarly, in the \texttt{RefCOCO} dataset, text queries typically appear in the form of descriptions such as "\textit{far right center of photo partial boat}". The "far right center" should be highlighted and emphasized in the reasoning process to improve performance. Nonetheless, LG-DVG regards the text query as a whole without sufficiently modeling the detailed semantic relation. Consequently, the performance of the LG-DVG model may be constrained due to its ineffectiveness in thoroughly capitalizing on the abundant linguistic information and associations present within the text query.

\section{Additional Experiments}

\subsection{Datasets}

\noindent \textbf{Flickr30k Entities}. The Flickr30k Entities dataset \cite{Flickr30kE} is a phrase localization dataset that extends the original Flickr30K \cite{Flickr30k} by including region-phrase correspondence annotations. It associates 31,783 images from Flickr30K \cite{Flickr30k} with 427K referred entities. Adopting the common splits employed in prior works \cite{VLTVG, MDGT, TransVG}, we allocate 29,783, 1,000, and 1,000 images for training, validation, and testing, respectively. This dataset presents two primary challenges: First, each image contains densely populated bounding boxes that may overlap with one another; second, a single phrase could correspond to multiple bounding boxes, resulting in a one-to-many relationship.

\noindent \textbf{RefCOCO-related}. Three referring expression datasets—RefCOCO \cite{RefCOCO}, RefCOCO+ \cite{RefCOCO}, and RefCOCOg \cite{RefCOCOg}—are based on images from the COCO dataset \cite{MSCOCO}. These datasets utilize natural language referring expressions to uniquely describe objects within images. RefCOCO \cite{RefCOCO} imposes no restrictions on the type of language used in referring expressions, while RefCOCO+ \cite{RefCOCO} constrains expressions to purely appearance-based descriptions, prohibiting location-based information in adherence to a computer vision perspective. In contrast, RefCOCOg \cite{RefCOCOg} contains more detailed object descriptions, as its annotations are collected in a non-interactive setting, resulting in an average of 8.4 words per expression compared to RefCOCO's 3.5 words.

RefCOCO \cite{RefCOCO} and RefCOCO+ \cite{RefCOCO} datasets comprise 19,994 and 19,992 images, respectively. Specifically, RefCOCO \cite{RefCOCO} includes 142,209 referring expressions for 50,000 objects, while RefCOCO+ \cite{RefCOCO} contains 141,564 expressions for 49,856 objects. Following the official splits of RefCOCO and RefCOCO+ \cite{RefCOCO}, the datasets are divided into train, val, testA, and testB subsets, with testA focusing on multiple persons and testB emphasizing multiple objects from other categories.

RefCOCOg \cite{RefCOCOg} comprises 25,799 images, which are accompanied by 49,856 reference objects and expressions. In our experiment, we employ the commonly used split protocol RefCOCOg-google \cite{Refcocogoogle}. 

\noindent \textbf{ReferItGame}. The ReferItGame dataset \cite{RIG} encompasses 130,525 referring expressions for 96,654 objects within 19,894 natural scene images, sourced from the SAIAPR-12 dataset \cite{SAIAPR12}. Following the configurations in \cite{TransVG}, we partition the samples into three subsets: The training set consists of 54,127 referring expressions, while the testing and validation sets contain 5,842 and 60,103 referring expressions, respectively.

\subsection{Learning}

\noindent \textbf{Inputs}. Following image augmentation mechanisms employed in prior research \cite{TransVG, MDGT}, we adjust the input image size to $640 \times 640$ by transforming the longer edge to $640$ and padding the shorter edge, while simultaneously preserving the original aspect ratio of the image.  In contrast to prior studies, such as \cite{TransVG, VLTVG}, which truncate the text query, our LG-DVG is capable of learning from text queries of variable lengths, demonstrating a more adaptive approach. In scenarios such as the \texttt{Flickr30k Entities} \cite{Flickr30kE} dataset, text queries in a batch consist of varying number of phrases. Thus, we first pad the number of phrases in these queries to the largest one of them by inserting empty phrases filled with [PAD] tokens. During this process, we generate the text mask, which is assigned $0$ to those empty phrases to exclude them from the computation of LG-DVG. After adding [CLS] and [SEP] tokens to each phrase using the Bert tokenizer, the phrases in a batch of text queries are padded to match the length of the longest phrase by adding [PAD] tokens.

\noindent \textbf{Learning settings}. A comprehensive summary of the detailed settings can be found in Table~\ref{table:Learning}, which presents an organized overview of the configuration information.

\noindent \textbf{Structure settings}. The structure settings of LG-DVG are shown in Table~\ref{table:Structure}.

\noindent \textbf{Supplementary methods}. In order to conduct a comprehensive comparison, we introduce three additional approaches, namely FAOA-iteration, FAOA-diffusion, and QRNet \cite{QRNet}. In particular, QRNet \cite{QRNet} represents an alternative method that emphasizes strengthening the visual backbone for subsequent visual grounding tasks. The FAOA-iteration approach constitutes an iterative version of FAOA \cite{FAOA}, wherein multiple iterations are executed, and the outputs from each preceding iteration serve as inputs for the following iteration. Meanwhile, FAOA-diffusion represents a diffusion variant of FAOA \cite{FAOA}, employing the FAOA \cite{FAOA} as a grounding decoder to process the noisy boxes at each Markov Step.

\noindent \textbf{Predictions}. Despite LG-DVG's reliance on box proposals for the learning process, it circumvents the use of Non-Maximum Suppression (NMS) in generating predictions by directly selecting predicted bounding boxes with the highest predicted similarity scores $\bm{\widehat{S}}$ as final outputs. In the case where $is\_ensemble$ is set to True in Algorithm \ref{algo:lgdvg-inference}, a variant version of LG-DVG, referred to as LG-DVG-nms, can be generated, with its final outputs obtained by applying Non-Maximum Suppression (NMS) to all $ensem\_boxes$.

\begin{table}[t]
\begin{center}
\caption{Detailed learning settings for Flickr30k Entities \cite{Flickr30kE} dataset. The source configuration file can be accessed under the \textit{configs/FK30} folder.}
\label{table:Learning}
\vspace{0.5cm}
\begin{adjustbox}{width=0.7\columnwidth}
\begin{tabular}{ll}
\hline
\multicolumn{1}{c|}{config}                      & value      \\ \hline
\multicolumn{1}{l|}{\#box proposals}             & 150        \\ 
\multicolumn{1}{l|}{\#epochs}                    & 60         \\ 
\multicolumn{1}{l|}{batch size}                  & 20         \\ \hline
\multicolumn{2}{|c|}{Optimizer}                                \\ \hline
\multicolumn{1}{l|}{name}                        & AdamW      \\ 
\multicolumn{1}{l|}{base learning rate}          & 1.0E-4     \\ 
\multicolumn{1}{l|}{weight decay}                & 1.0E-4     \\ 
\multicolumn{1}{l|}{betas}                        & 0.9, 0.999 \\ \hline
\multicolumn{2}{|c|}{LR scheduler}                             \\ \hline
\multicolumn{1}{l|}{name}                        & Cosine     \\ 
\multicolumn{1}{l|}{lr backbone}                 & 1.0E-4     \\ 
\multicolumn{1}{l|}{warm lr}                     & 1.0E-6     \\ 
\multicolumn{1}{l|}{min lr}                      & 1.0E-7     \\ 
\multicolumn{1}{l|}{warmup epochs}               & 5          \\ 
\multicolumn{1}{l|}{cooldown epochs}             & 5          \\ 
\multicolumn{1}{l|}{decay\_rate}                 & 0.1        \\ 
\multicolumn{1}{l|}{lr\_noise}                   & 0.001      \\ 
\multicolumn{1}{l|}{lr\_noise\_percent}          & 0.67       \\ 
\multicolumn{1}{l|}{lr\_noise\_std}              & 1.0        \\ 
\multicolumn{1}{l|}{lr\_cycle\_limit}            & 1          \\ 
\multicolumn{1}{l|}{lr\_cycle\_mul}              & 1          \\ 
\multicolumn{1}{l|}{gradient accumulation steps} & 1          \\ 
\multicolumn{1}{l|}{clip max norm}               & 0.1        \\ \hline
\multicolumn{2}{|c|}{Losses}                             \\ \hline
\multicolumn{1}{l|}{$\alpha$ for $\mathcal{L}_{smooth-l1}$} & 2    \\     
\multicolumn{1}{l|}{$\beta$ for $\mathcal{L}_{giou}$} & 5 \\ 
\multicolumn{1}{l|}{$\lambda$ for $\mathcal{L}_{\bm{\widehat{S}}}$} & 1 \\ 
\hline

\end{tabular}
\end{adjustbox}
\end{center}
\vspace{-0.7cm}
\end{table}

\begin{table}[t]
\begin{center}
\caption{Detailed structure settings of LG-DVG. The source configuration file can be accessed under the \textit{configs} folder.}
\label{table:Structure}
\vspace{0.6cm}
\begin{adjustbox}{width=0.8\columnwidth}
\begin{tabular}{ll}
\hline
\multicolumn{1}{l|}{config}                     & value                 \\ \hline
\multicolumn{1}{l|}{box proposals}              & 150                   \\ \hline
\multicolumn{1}{l|}{image size}                 & $640 \times 640$               \\ \hline
\multicolumn{2}{|c|}{Diffusion}                                          \\ \hline
\multicolumn{1}{l|}{chain length}               & 1000                  \\ \hline
\multicolumn{1}{l|}{variance scheduler $\beta$} & cosine scheduler      \\ \hline
\multicolumn{1}{l|}{cosine s}                   & 0.008                 \\ \hline
\multicolumn{1}{l|}{scale}                      & 2.0                   \\ \hline
\multicolumn{2}{|c|}{Structure}                                          \\ \hline
\multicolumn{1}{l|}{visual encoder}             & out: 256              \\ \hline
\multicolumn{1}{l|}{text encoder}               & out: 768              \\ \hline
\multicolumn{1}{l|}{text encoder pretrained}               &  whaleloops/phrase-bert              \\ \hline

\multicolumn{1}{l|}{text projection}            & in: 768, out: 256      \\ \hline
\multicolumn{1}{l|}{scale\_shift\_mapper}       & in: 512, out:512      \\ \hline
\multicolumn{2}{|c|}{Rois}                                               \\ \hline
\multicolumn{1}{l|}{pooler type}                & MultiScaleRoIAlign    \\ \hline
\multicolumn{1}{l|}{pooler resolution}          & 7                     \\ \hline
\multicolumn{1}{l|}{pooler sampling ratio}      & 2                     \\ \hline
\multicolumn{1}{l|}{box\_projection}            & hidden: 512, out: 256 \\ \hline
\multicolumn{2}{|c|}{cross-modal transformer}                            \\ \hline
\multicolumn{1}{l|}{number of blocks}           & 1                     \\ \hline
\multicolumn{1}{l|}{number of heads}            & 8                     \\ \hline
\multicolumn{1}{l|}{qkv features}               & 256                   \\ \hline
\multicolumn{1}{l|}{projection features}        & 256                   \\ \hline
\multicolumn{1}{l|}{attention dropout}          & 0.0                   \\ \hline
\multicolumn{1}{l|}{projection dropout}         & 0.0                   \\ \hline
\multicolumn{1}{l|}{feedforward}                & hidden: 512, out: 256 \\ \hline
\multicolumn{2}{|c|}{bbox\_regression}                                   \\ \hline
\multicolumn{1}{l|}{number of layers}           & 1                     \\ \hline
\multicolumn{1}{l|}{in features}                & 256                   \\ \hline
\multicolumn{1}{l|}{out feature}                & 256                   \\ \hline
\multicolumn{1}{l|}{prediction}                 & 4                     \\ \hline
\multicolumn{1}{l|}{box coordinates weight}     & 2.0, 2.0, 1.0, 1.0    \\ \hline

\end{tabular}
\end{adjustbox}
\end{center}
\vspace{-0.6cm}
\end{table}

\noindent \textbf{Reproducibility}. All experiments involving our proposed LG-DVG model are conducted using the publicly accessible platform, \texttt{VGBase}. The source code can be found in the \textit{code/LGDiffusionVG} folder, while the majority of configuration files are located within \textit{code/configs}. We offer comprehensive descriptions for each Python file under \textit{code/LGDiffusionVG} and provide instructions on executing the code to replicate the results presented in our paper. Lastly, as indicated in our configuration file, the random $seed$ is set to $6$ to ensure reproducibility.

\subsection{Supplementary Quantitative Results}

Table~\ref{tab:fk30sota} presents a performance comparison between our proposed LG-DVG method and all possible state-of-the-art approaches with diverse structures. By incorporating the ensemble mechanism for LG-DVG, its $Acc@0.5$ becomes the second approach to surpass the $0.8$ threshold, achieving a score of $80.77$. Maintaining the same inference time, this outcome surpasses the top-performing iterative visual grounding method, M-DGT, by $0.8$, and exhibits an improvement of $0.85$ over the original LG-DVG. We deduce that the improvement stems from the ensemble mechanism incorporating all predicted boxes during the reverse diffusion process to match with ground truth for evaluation. In contrast, the original LG-DVG method selects the final predictions based on the predicted similarity scores, $\bm{\widehat{S}}$, which may potentially contain deviations, thereby affecting the overall performance. Nonetheless, to ensure a fair comparison, the primary focus of this paper does not encompass the ensemble version of LG-DVG, since the alternative methods do not employ the ensemble mechanism.

\begin{table}
\begin{center}
\caption{Comparison of phrase localization performance with state-of-the-art methods using the metric $Acc@0.5$ on the test sets of \texttt{Flickr30k Entities}(Flickr30kE) \cite{Flickr30kE} and \texttt{ReferItGame}/\texttt{RefCLEF} \cite{RIG}. The best performance is emphasized using bold black font.}
\label{tab:fk30sota}
\vspace{0.3cm}
\begin{adjustbox}{max width=\columnwidth}
\begin{tabular}{lcccc}
\toprule
Method & \begin{tabular}[c]{@{}c@{}}Visual \\ Encoder\end{tabular} & \begin{tabular}[c]{@{}c@{}}Text \\ Encoder\end{tabular} & \texttt{Flickr30kE} & \texttt{RefCLEF}  \\

\hline
\multicolumn{5}{c}{\textit{State-of-the-art approaches}}  \\ 
\hline

FAOA \cite{FAOA}                        &   DN53           &  Bert             &  68.69    &  60.67        \\
FAOA+RED \cite{DeconVG}                 &   DN53           &  Bert             &  70.50    &  67.27        \\
FAOA+iteration                 &   DN53           &  Bert             &  67.87    &  -        \\
FAOA+diffusion                 &   DN53           &  Bert             &  74.16    &  -        \\
TransVG \cite{TransVG}                  &   RN101          &  Bert             &  79.10    &  70.73     \\
TransVG \cite{TransVG}                  &   RN101          &  Phr-Bert         &  $\times$        &  $\times$     \\
TransVG \cite{TransVG}                  &   SwimS          &  Bert             &  78.18        &  70.86     \\
ReSC-large  \cite{RecursiveSubQuery}    &   DN53            &  Bert             &  69.28    &  64.6      \\
VLTVG \cite{VLTVG}                      &   RN50            &  Bert             &  79.18    &  71.60      \\
VLTVG \cite{VLTVG}                      &   RN101           &  Bert             &  79.84    &  71.98       \\
VLTVG \cite{VLTVG}                      &   RN101           &  Phr-Bert         &  $\times$    &  $\times$       \\
M-DGT \cite{MDGT}                       &   RN50            &  Bert             &  79.32    &  72.41     \\
M-DGT \cite{MDGT}                       &   RN101           &  Bert             &  79.97    &  73.63     \\
QRNet \cite{QRNet}                      &   SwimS           &  Bert             &  \textbf{81.95}    &  \textbf{74.61}     \\

\hline

\hline
\multicolumn{5}{c}{\textit{LG-DVG}}  \\ 
\hline
LG-DVG                                  &   RN101           &  Phr-Bert      &  79.21    &  71.46          \\
LG-DVG                                  &   SwinB           &  Phr-Bert      &  79.92    &  72.82           \\

\hline
\multicolumn{5}{c}{\textit{LG-DVG} with ensemble as shown in Algorithm~\ref{algo:lgdvg-inference}}  \\ 
\hline

LG-DVG-nms                                  &   RN101           &  Phr-Bert      &  80.4    &  74.02          \\
LG-DVG-nms                                  &   SwinB           &  Phr-Bert      &  80.77    &  74.24           \\

\hline

\end{tabular}
\end{adjustbox}
\end{center}
\vspace{-0.4cm}
\end{table}

\begin{figure}[t]
    \centering
    \includegraphics[width=\columnwidth]{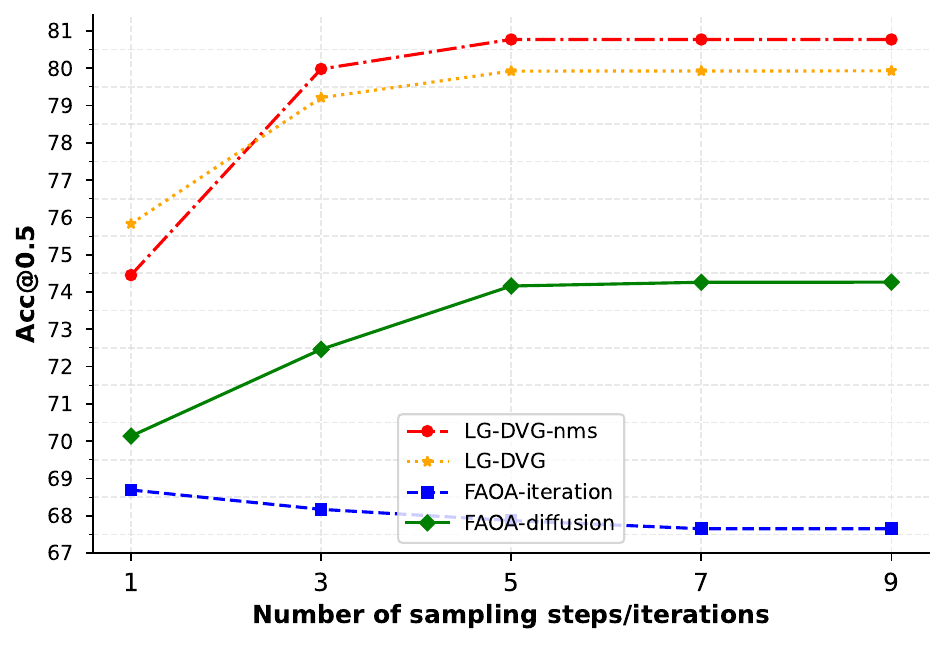}
    \caption{A comparison conducted between the original LG-DVG, LG-DVG with ensemble (LG-DVG-nms), FAOA-iteration, and FAOA-diffusion on the test set of \texttt{Flickr30k Entities} \cite{Flickr30kE}. The objective of this comparison is to demonstrate the progressive refinement properties of various approaches under differing numbers of sampling steps or iterations.}
    \label{fig:morecompare}
    \vspace{0.1cm}
\end{figure}

Figure~\ref{fig:morecompare} illustrates the performance of these approaches as the number of sampling steps or iterations increases. LG-DVG-nms rapidly achieves a higher Acc@0.5 within five sampling steps, as all predicted bounding boxes are combined using NMS to generate the final predictions, which are bound to include the most accurate ones. Direct evidence of this improvement can also be found in Table~\ref{tab:ablationensemble}. However, in the ablation study where we remove the similarity loss $\mathcal{L}_{\bm{\widehat{S}}}$, LG-DVG-nms does not yield significant improvement; in fact, Acc@0.5 declines to $77.372$. Moreover, the performance decreases in terms of Acc@0.5 and Acc@0.7 when removing $\mathcal{L}_{\bm{\widehat{S}}}$ during training, highlighting the importance of incorporating box-phrase similarity in the learning process. We contend that our language-guided approach necessitates learning semantic correspondences between queries and bounding boxes, enabling each box to be guided by a specific query to adjust its coordinates.

\begin{table}
\begin{center}
\caption{Ablation studies of LG-DVG in \texttt{Flickr30k Entities} \cite{Flickr30kE} dataset. The best setup is emphasized using the gray background. The term "Ensemble" refers to the condition where the $is_ensemble$ parameter in Algorithm \ref{algo:lgdvg-inference} is set to be True. Meanwhile, $\mathcal{L}_{\bm{\widehat{S}}}$ denotes the inclusion of similarity loss during the training process.}
\label{tab:ablationensemble}
\begin{adjustbox}{max width=\columnwidth}
\begin{tabular}{c|c|c|c|c|c}
\hline

\hline

    Ensemble                   & \multicolumn{2}{c|}{Similarity loss $\mathcal{L}_{\bm{\widehat{S}}}$}          & Acc@0.5          & Acc@0.7      & Sampling Steps    \\ 
\hline
                           & \multicolumn{2}{c|}{}    & 76.27            & 68.481      & 8                \\
                           & \multicolumn{2}{c|}{\checkmark}    & 79.924            & 70.151      & 5                \\
\checkmark                 & \multicolumn{2}{c|}{}              & 77.372            & 68.945      & 4                  \\
\rowcolor{gray!25} \checkmark                 & \multicolumn{2}{c|}{\checkmark}    & 80.772            & 72.34      & 5                  \\                           
\hline
\hline
\end{tabular}
\end{adjustbox}
\end{center}
\vspace{0.1cm}
\end{table}

\section{One-to-many Challenge}
\begin{figure}[ht]
    \centering
    \includegraphics[width=\columnwidth]{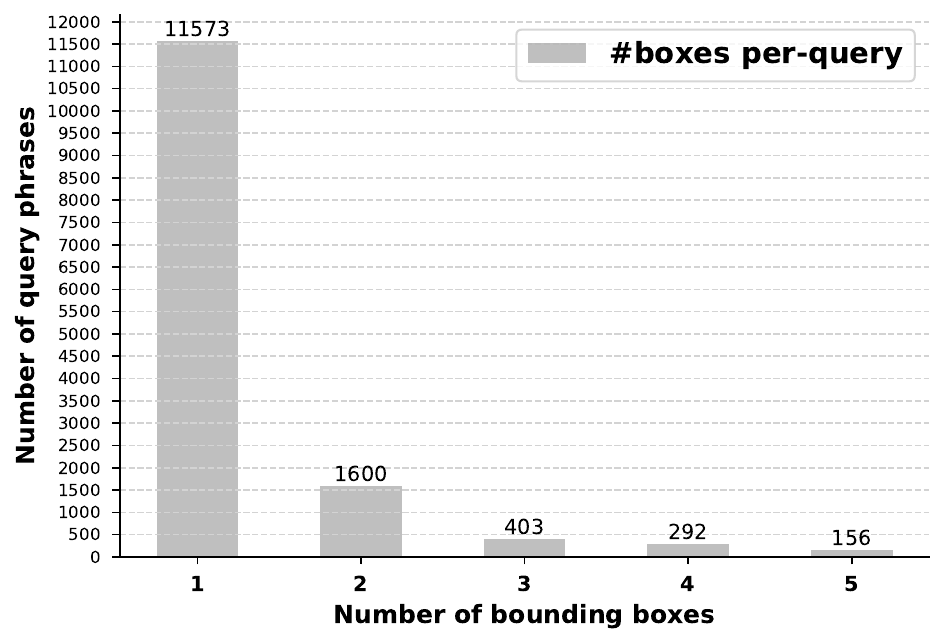}
    \includegraphics[width=\columnwidth]{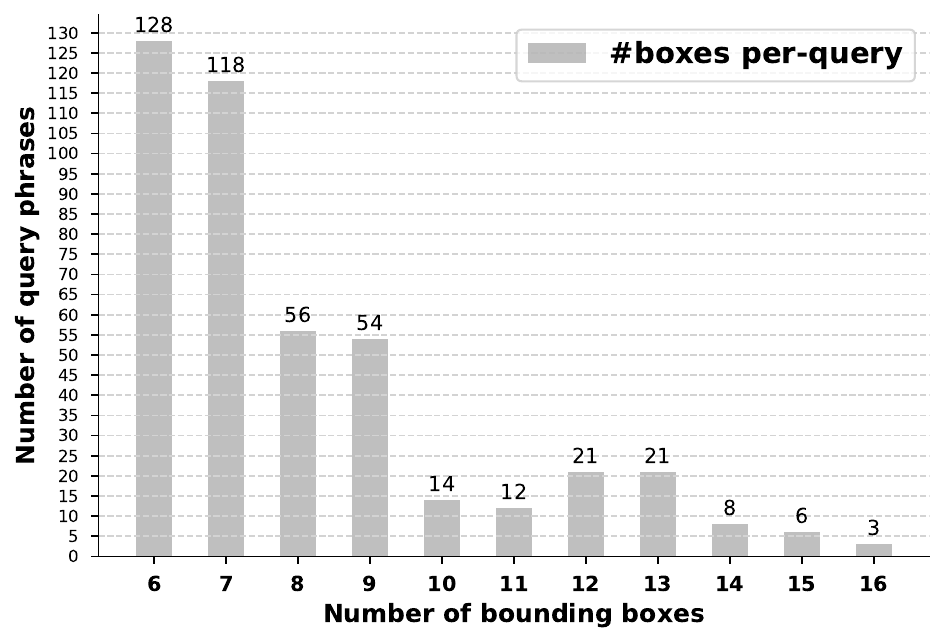}
    \caption{One-to-many statistics on the test set of the \texttt{Flickr30k Entities} \cite{Flickr30kE}.}
    \label{fig:onetomanysta}
    \vspace{-0.1cm}
\end{figure}

\begin{figure}[ht]
    \centering
    \includegraphics[width=\columnwidth]{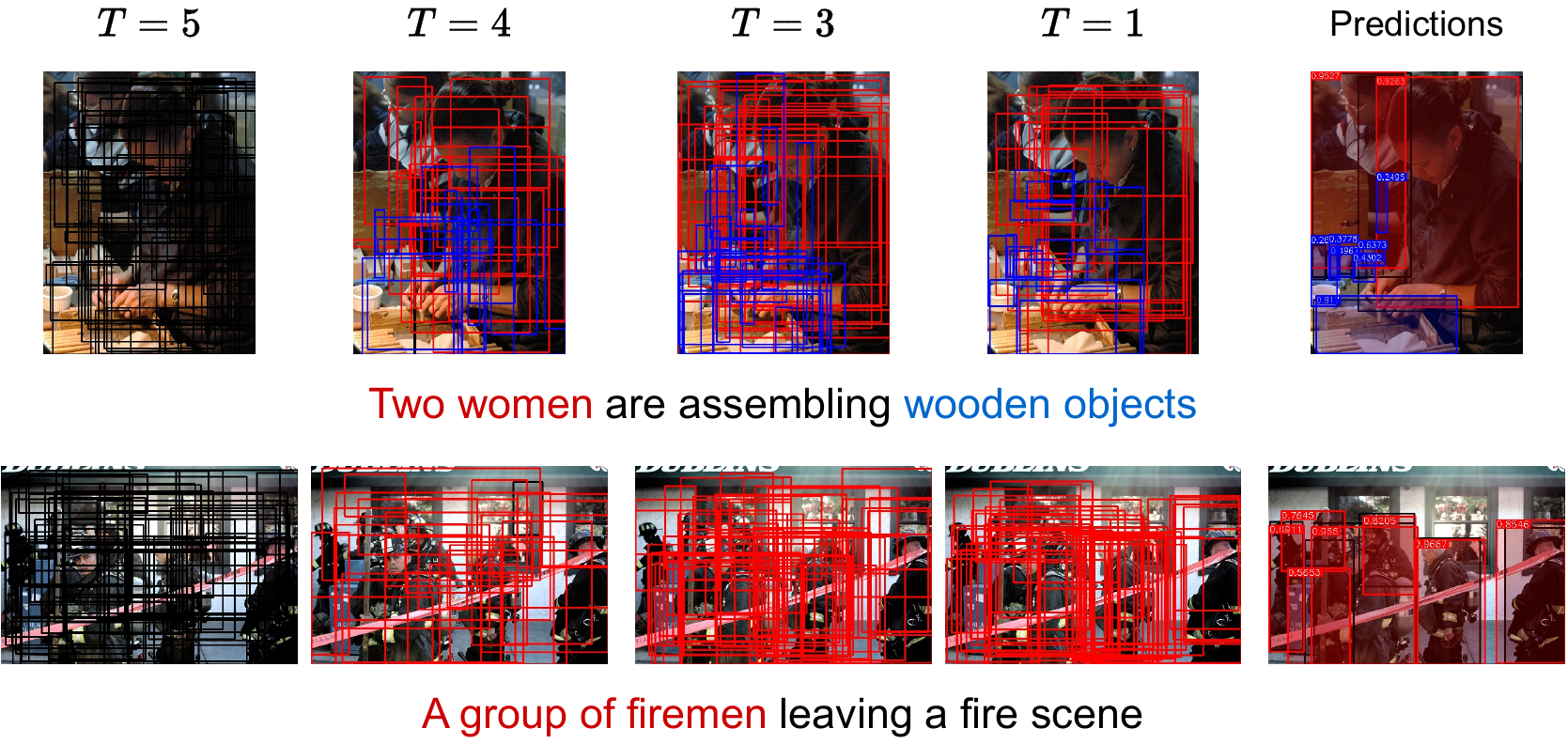}
    \caption{A single query is associated with $\bm{7}$ distinct regions within the image.}
    \label{fig:onetoseven}
    \vspace{-0.1cm}
\end{figure}

\begin{figure}[ht]
    \centering
    \includegraphics[width=\columnwidth]{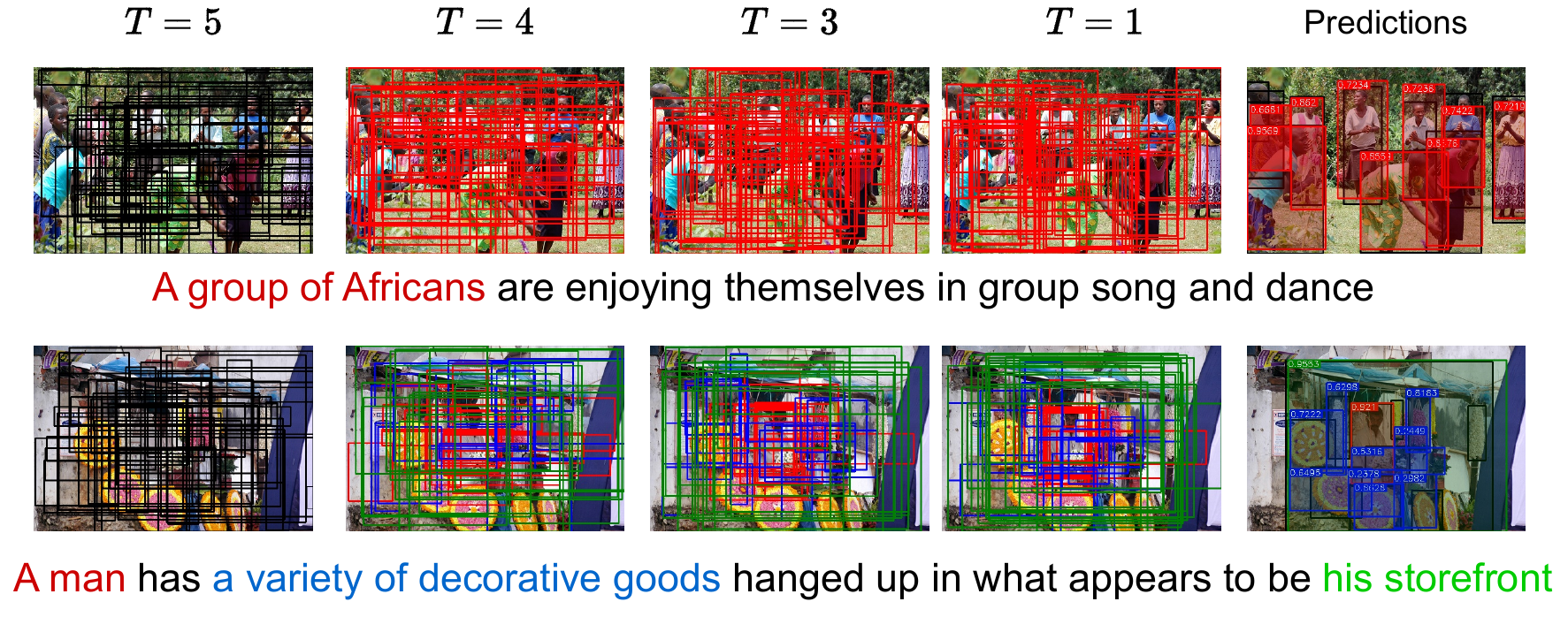}
    \caption{A single query is associated with $\bm{9}$ distinct regions within the image.}
    \label{fig:onetonine}
    \vspace{-0.1cm}
\end{figure}

\begin{figure}[ht]
    \centering
    \includegraphics[width=\columnwidth]{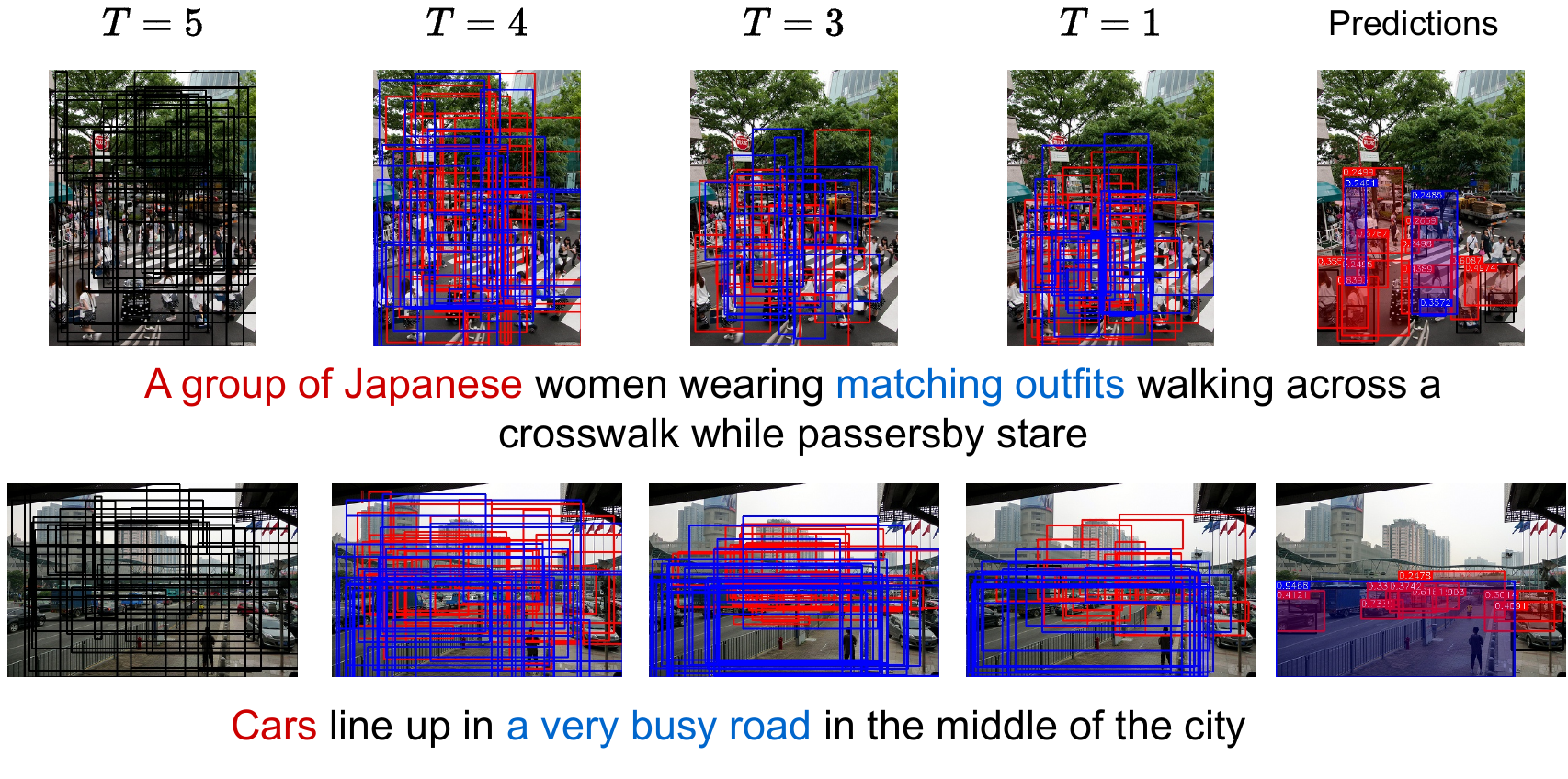}
    \caption{A single query is associated with $\bm{11}$ distinct regions within the image.}
    \label{fig:onetoeleven}
    \vspace{-0.1cm}
\end{figure}

\begin{figure}[t]
    \centering
    \includegraphics[width=\columnwidth]{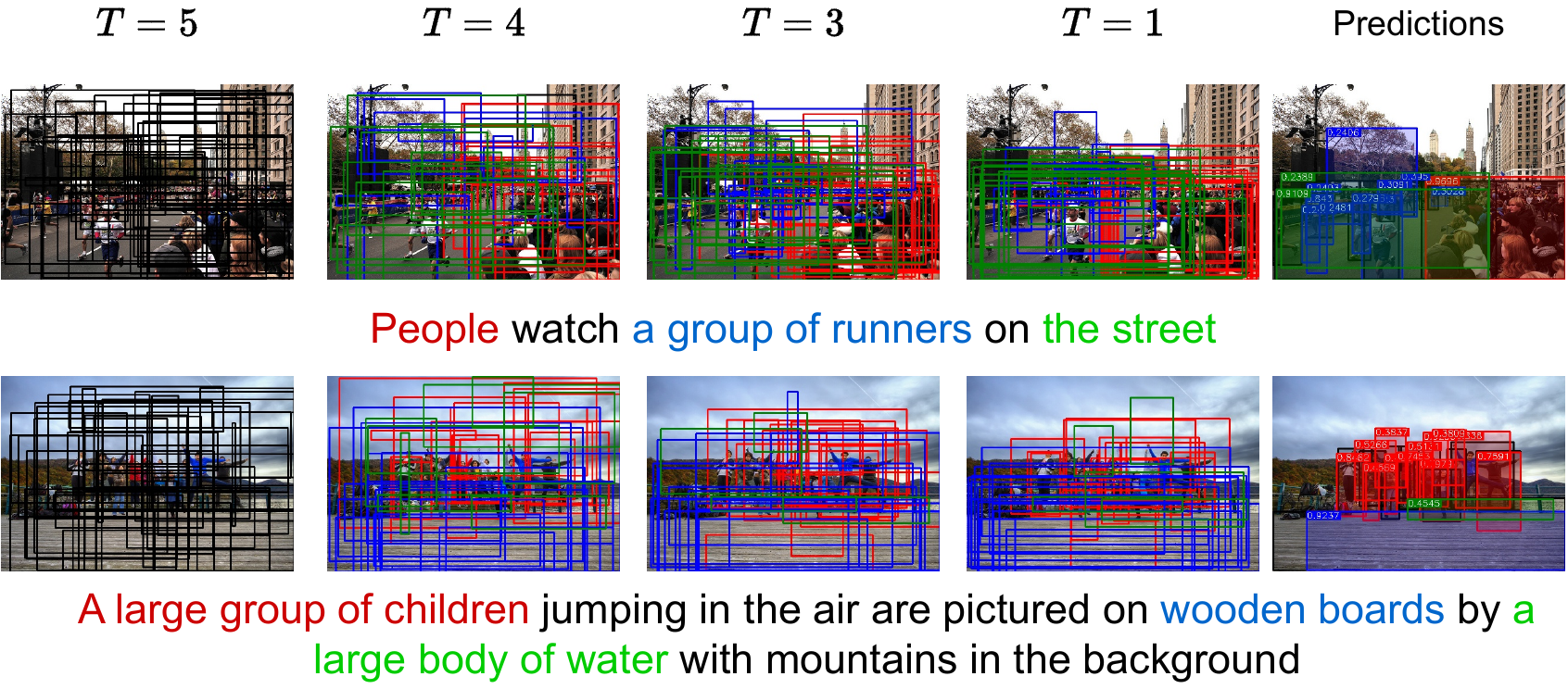}
    \caption{A single query is associated with $\bm{13}$ distinct regions within the image.}
    \label{fig:oneto13}
    \vspace{-0.1cm}
\end{figure}

\begin{figure*}[ht]
    \centering
    \includegraphics[width=0.95\textwidth]{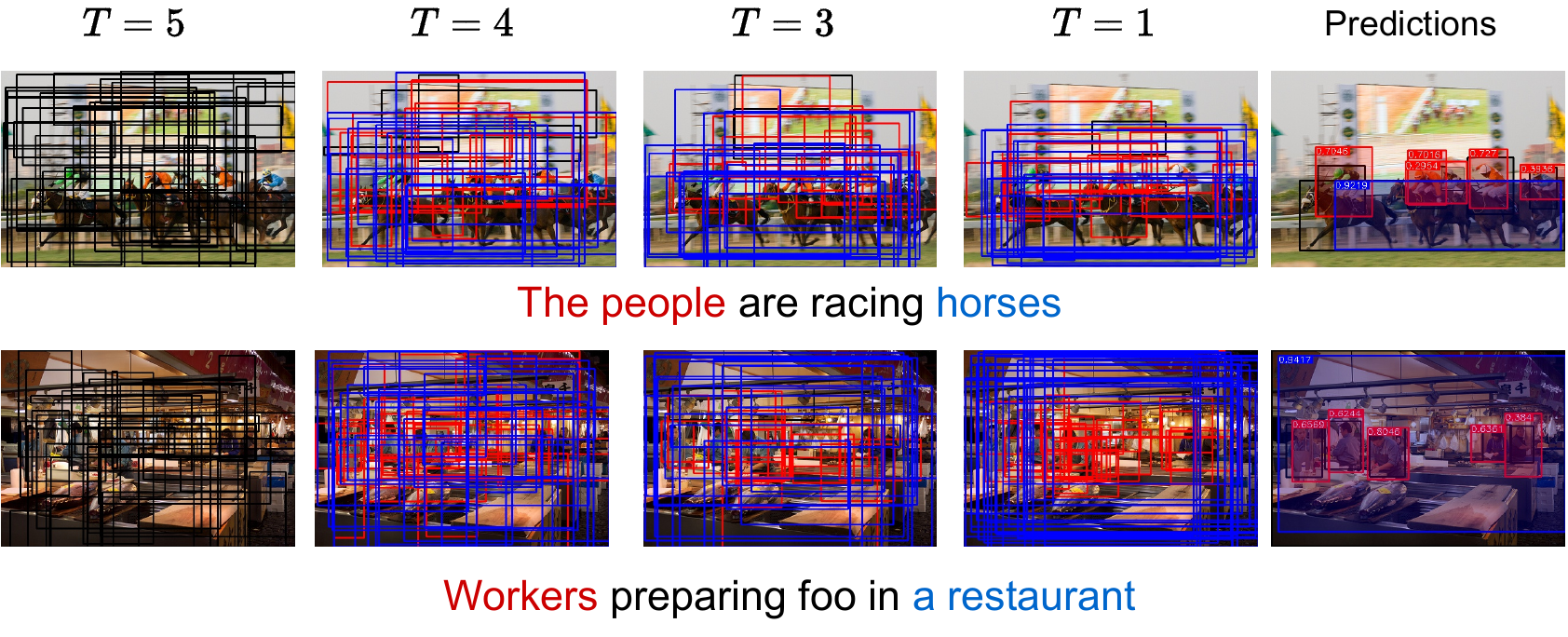}
    \caption{Inference outputs of LG-DVG on the test set of \texttt{Flickr30k Entities} data \cite{Flickr30kE}. A single query is associated with $\bm{5}$ distinct regions within the image. The first column $T=5$ is the first denoising step from randomly generated boxes. The final column, labeled $Predictions$ displays the ultimate predicted boxes, with their colors matching those of the associated text queries. The obtained $\bm{\widehat{S}}$ is presented at the upper left corner of each bounding box.} 
    \label{fig:onetofive}
    \vspace{-0.5cm}
\end{figure*}

\begin{figure}[ht]
    \centering
    \includegraphics[width=\columnwidth]{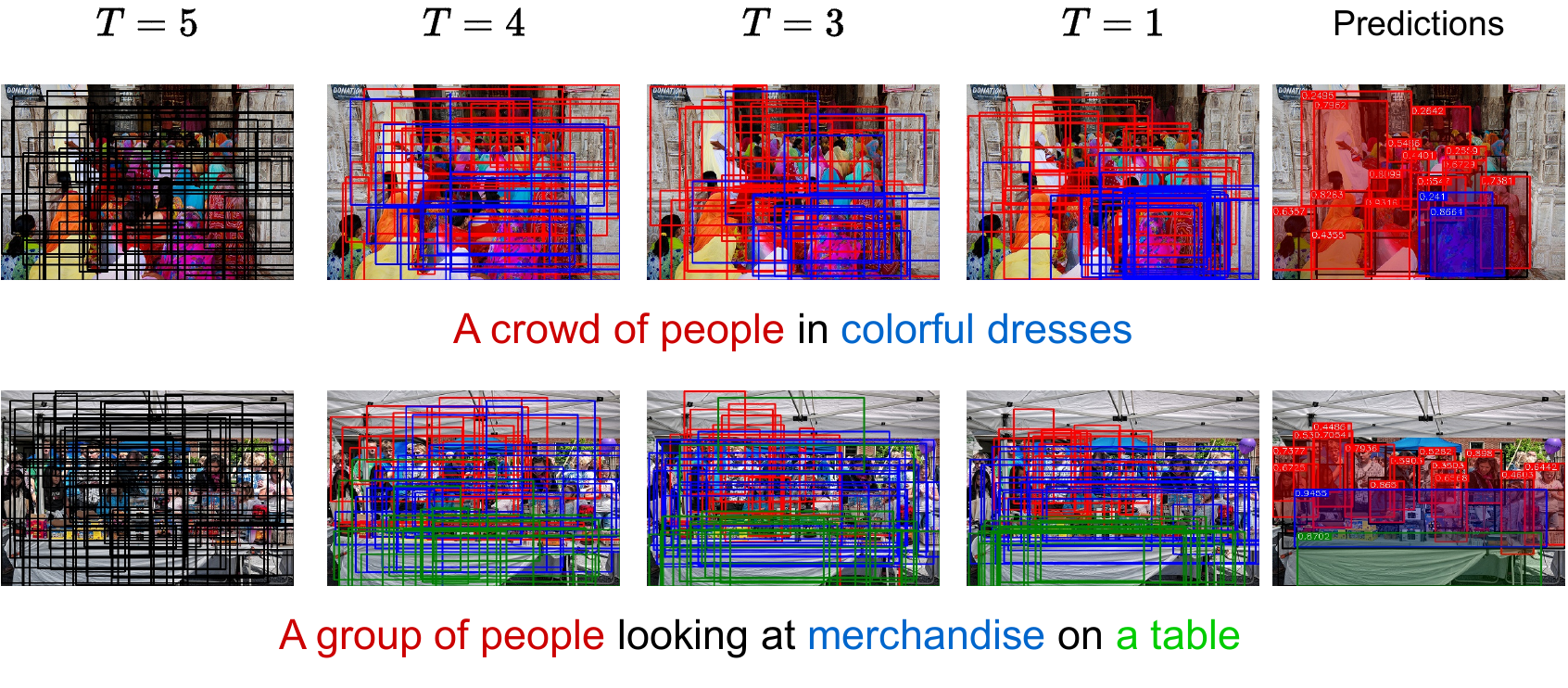}
    \caption{A single query is associated with $\bm{15}$ distinct regions within the image.}
    \label{fig:oneto15}
    \vspace{-0.1cm}
\end{figure}

The statistic of the one-to-many cases in the test set of the Flick30K dataset is illustrated in Figure~\ref{fig:onetomanysta}. Despite the fact that precisely 11,575 queries are exclusively associated with a single bounding box, it is crucial to underscore the significance of addressing the one-to-many challenge. This is primarily because advanced visual grounding methodology should follow human perception, which enables the detection of all potential objects corresponding to a text query. Our proposed LG-DVG presents a solid ability to locate all possible regions semantically aligned with the text query. Additional qualitative outcomes, derived from the evaluation of the test dataset within the \texttt{Flickr30k Entities} data \cite{Flickr30kE}, are presented in Figures~\ref{fig:onetofive}, \ref{fig:onetoseven}, \ref{fig:onetonine}, \ref{fig:onetoeleven}, \ref{fig:oneto13}, \ref{fig:oneto15}. The presented visual outcomes illustrate the successful performance of our LG-DVG in achieving one-to-many matching across various scenarios, where a single query may align with $\bm{5}$, $\bm{7}$, $\bm{9}$, $\bm{11}$, $\bm{13}$, or even $\bm{15}$ bounding boxes.

\section{Tight Bounding Boxes}

\begin{figure}[t]
    \centering
    \includegraphics[width=\columnwidth]{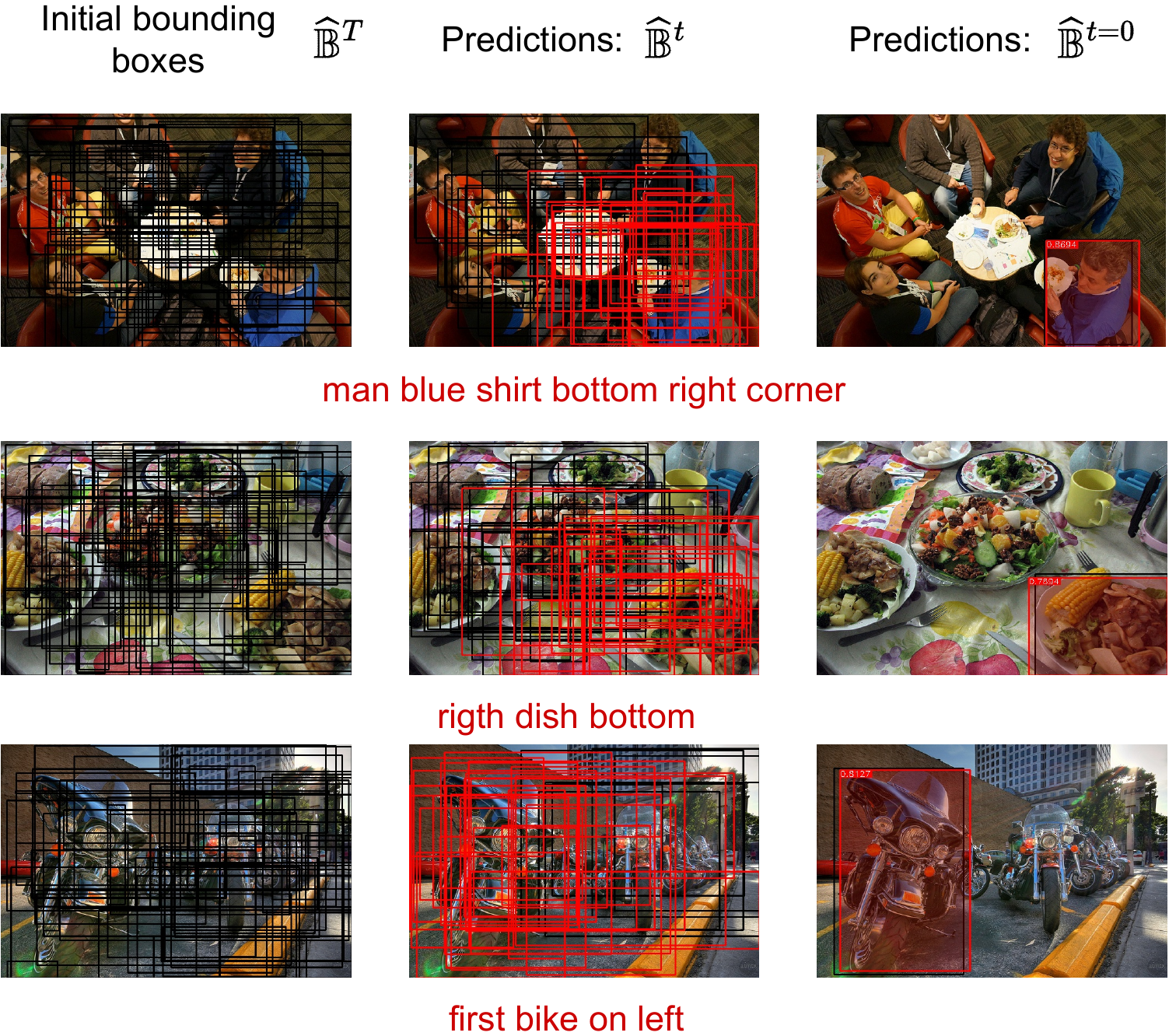}
    \caption{Inference outputs of LG-DVG on the test set of \texttt{RefCOCO} data \cite{RefCOCO} with \textit{google} split. We present visualizations of the sampling step in inference.}
    \label{fig:refcoco}
    \vspace{-0.1cm}
\end{figure}

\begin{figure}[t]
    \centering
    \includegraphics[width=0.9\columnwidth]{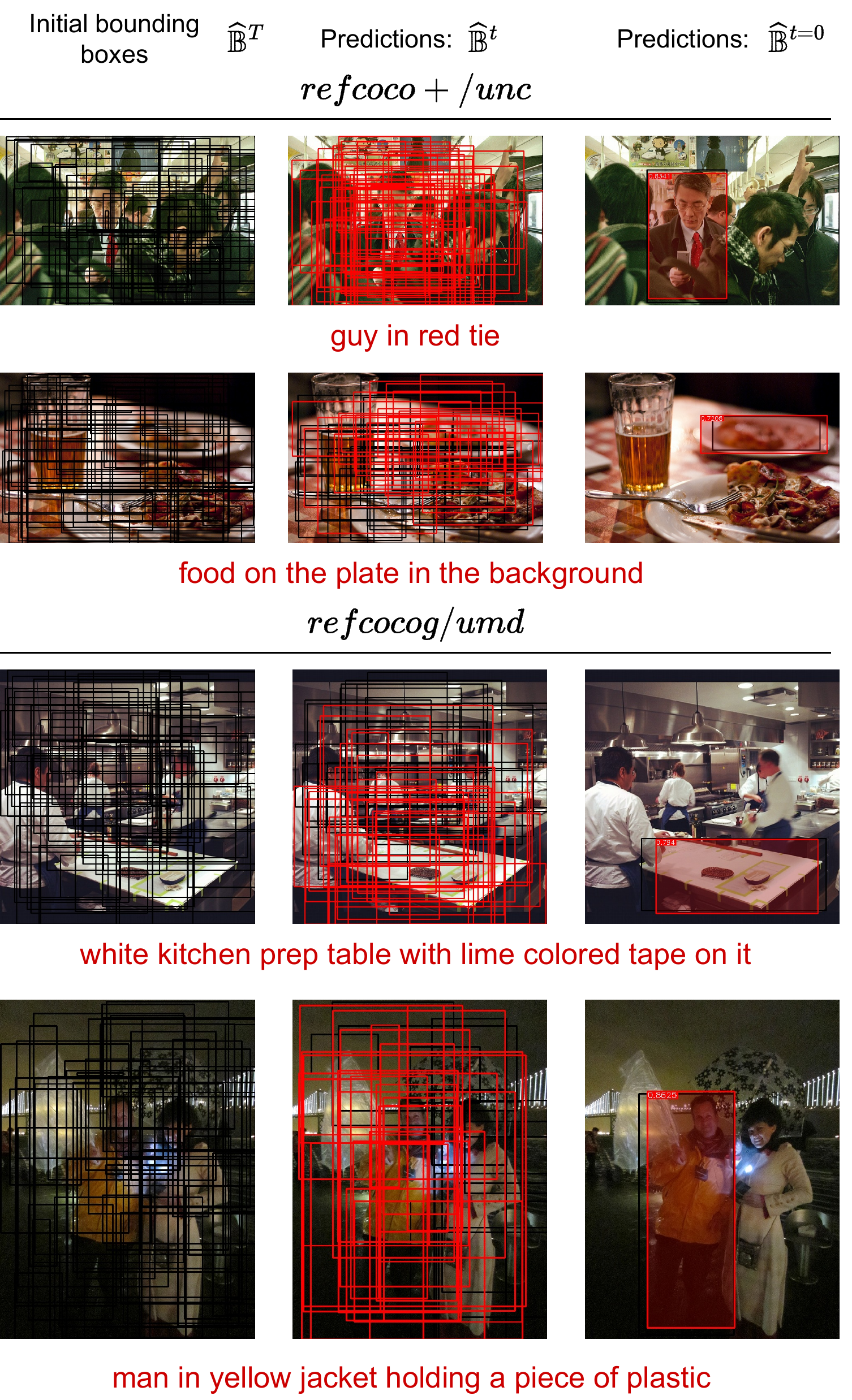}
    \caption{Inference outputs of LG-DVG on the test set of \texttt{RefCOCO+} data \cite{RefCOCO} and \texttt{RefCOCOg} with \textit{unc} and \textit{umd} splits, respectively. We present visualizations of the sampling step in inference.}
    \label{fig:refcocog+}
    \vspace{-0.1cm}
\end{figure}

Owing to the potent attribute of progressive refinement, LG-DVG can generate tight bounding boxes that exhibit high Intersection over Union (IOU) scores between each box and ground truth boxes. In addition to the quantitative findings presented in the primary body of the paper, we offer supplementary qualitative results in Figures~\ref{fig:refcoco}, \ref{fig:refcocog+} to explicitly demonstrate the degree of tightness achieved by the bounding boxes generated through the LG-DVG method.

\end{document}